\documentclass[10pt,twocolumn,letterpaper]{article}

\usepackage[pagenumbers]{cvpr} %
\usepackage{graphicx}
\usepackage{amsmath}
\usepackage{amssymb}
\usepackage{booktabs}
\usepackage{comment}

\usepackage[pagebackref,breaklinks,colorlinks]{hyperref}
\usepackage[mathscr]{euscript}
\usepackage{tabularx}

\usepackage[capitalize]{cleveref}
\crefname{section}{Sec.}{Secs.}
\Crefname{section}{Section}{Sections}
\Crefname{table}{Table}{Tables}
\crefname{table}{Tab.}{Tabs.}

\newcommand{\z}{\mathbf{z}}
\newcommand{\zglob}{\mathbf{z}_{\text{glob}}}

\newcommand{\zid}{\mathbf{z}^{\text{id}}}
\newcommand{\zidglob}{\mathbf{z}_{\text{glob}}^{\text{id}}}

\newcommand{\zex}{\mathbf{z}^{\text{ex}}}

\newcommand{\numScans}{5200~}
\newcommand{\numIDs}{255~}

\newcommand{\OURS}{NPHM}

\DeclareMathOperator*{\argmin}{arg\,min}

\def\confName{CVPR}
\def\confYear{2023}

\begin{document}

\title{Learning Neural Parametric Head Models}

\author{
Simon Giebenhain$^1$ \quad
Tobias Kirschstein$^1$ \quad
Markos Georgopoulos$^2$ \quad
Martin Rünz$^2$ \\
Lourdes Agapito$^3$ \quad
Matthias Nie{\ss}ner$^1$ \vspace{0.2cm}\\
$^1$Technical University of Munich  \qquad $^2$Synthesia \qquad $^3$University College London
}

\twocolumn[{
\renewcommand\twocolumn[1][]{#1}
\maketitle
\thispagestyle{empty}
\begin{center}
  \vspace{-0.3cm}
  \newcommand{\teaserwidth}{\textwidth}
  \centerline{
    \includegraphics[width=\teaserwidth]{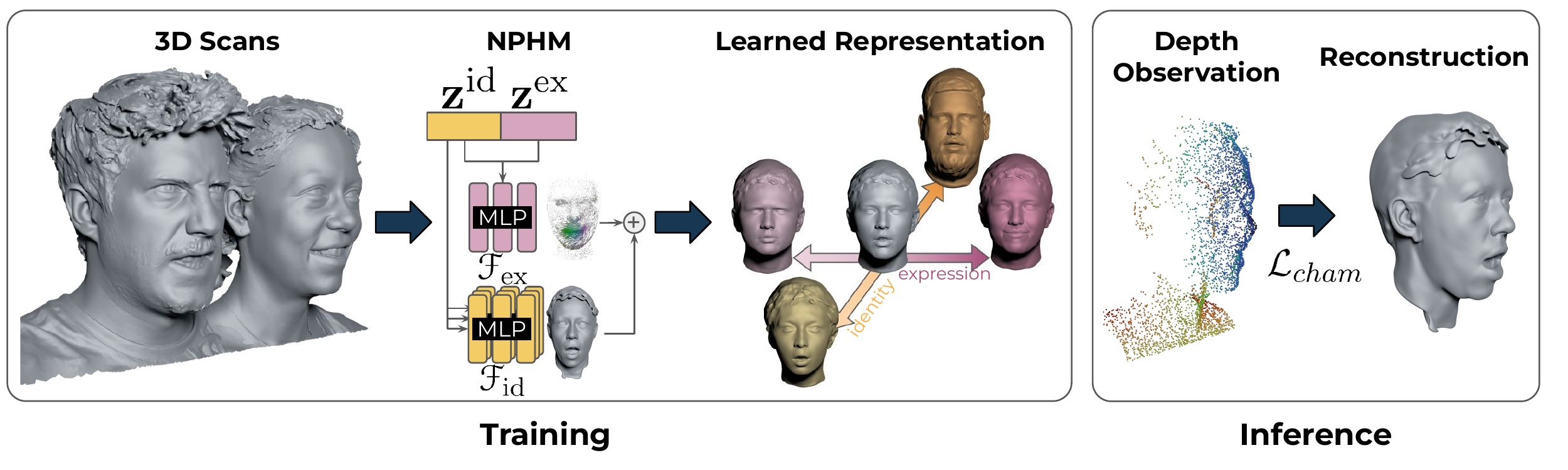}
    \vspace{-0.3cm}
    }
    \captionof{figure}{We propose to learn a neural parametric head model based on neural fields: first, we capture a large dataset of over \numScans high-fidelity head scans with varying shapes and expressions (left). We then non-rigidly register these scans to generate our training data. As a result of training, we obtain a disentangled latent that spans the space of shapes $\zid$ and expressions $\zex$ (middle). At inference time, we can leverage the prior of our learned representation by fitting our model to a sparse input point cloud by solving for the latent codes (right).
    }
    \vspace{-0.15cm}
  \label{fig:teaser}
 \end{center}
}]

\begin{abstract}
We propose a novel 3D morphable model for complete human heads based on hybrid neural fields.
At the core of our model lies a neural parametric representation that disentangles identity and expressions in disjoint latent spaces.
To this end, we capture a person's identity in a canonical space as a signed distance field (SDF), and model facial expressions with a neural deformation field.
In addition, our representation achieves high-fidelity local detail by introducing an ensemble of local fields centered around facial anchor points. %
To facilitate generalization, we train our model on a newly-captured dataset of over \numScans head scans from \numIDs different identities using a custom high-end 3D scanning setup.
Our dataset significantly exceeds comparable existing datasets, both with respect to quality and completeness of geometry, averaging around 3.5M mesh faces per scan.
Finally, we demonstrate that our approach outperforms state-of-the-art methods in terms of fitting error and reconstruction quality. %
{\let\thefootnote\relax\footnotetext{\footnotesize{Website: \url{https://simongiebenhain.github.io/NPHM}}}}
\end{abstract}

\section{Introduction}

Human faces and heads lie at the core of human visual perception, and hence are key to creating digital replica of someones identity, likeliness, and appearance. 
In particular, 3D reconstruction of human heads from sparse inputs, such as point clouds, is central to a wide range of applications in the context of gaming, augmented and virtual reality, and digitization in our modern digital era.
One of the most successful lines of research to address this challenging problem are parametric face models, which represent both shape identities and expressions featuring a low-dimensional parametric space.
These Blendshape and 3D morphable models (3DMMs) have achieved incredible success, since they can be fitted to sparse inputs, regularize out noise, and provide a compact 3D representation.
As a result, many practical settings could be realized, ranging from face tracking and 3D avatar creation to facial-reenactment applications~\cite{zollhofer2018state}.

Traditionally, 3DMMs, are based on a low-rank approximation of the underlying 3D mesh geometry.
To this end, a template mesh with fixed topology is non-rigidly registered to a series of 3D scans. 
From this template registration, a 3DMM can be computed using dimensionality reduction methods such as principal component analysis (PCA).
The quality of the resulting parametric space depends strongly on the quality of 3D scans, their registration, and the ability to disentangle identity and expression variations.
While these PCA-based models exhibit excellent regularizing properties, their inherent limitation lies in their inability to represent local surface detail and the reliance on a template mesh of fixed topology, which inhibits the representation of diverse hair styles.

In this work, we propose neural parametric head models (\OURS), which represent complete human head geometry in a canonical space using an SDF, and morph the resulting geometry to posed space using a forward deformation field. 
By decoupling the human head representation into these two spaces, we are able to learn disentangled latent spaces -- one of the core concepts of 3DMMs.
Furthermore, we decompose the implicit geometry representation in canonical space into an ensemble of local MLPs.
Each part is represented by a small MLP that operates in a local coordinate system centered around face keypoints. 
Additionally, we exploit face symmetry by sharing network weights of symmetric regions. 
This decomposition into separate parts imposes a strong geometry prior and helps to improve both generalization and provide higher levels of detail.

In order to train our model, we capture a new high-fidelity head dataset with a high-end capture rig, which is composed of over \numScans 3D head scans from \numIDs different people.
After rigidly aligning all scans in a canonical coordinate system, we train our identity network on scans in canonical expression.
In order to train the deformation network, we non-rigidly register each scan against a template mesh, which we in turn use as training data for our neural deformation model.
At inference time, we can then fit our model to a given input point cloud by optimizing for the latent code parameters for both expression and identity. 
In a series of experiments, we demonstrate that our neural parametric model outperforms state-of-the-art models and can represent complete heads, including fine details.\\
\smallskip\noindent
In sum, our contributions are as follows:
\vspace{-0.1cm}
\begin{itemize}
\setlength\itemsep{-.3em}
    \item We introduce a novel 3D dataset captured with a high-end capture rig, including over \numScans 3D scans of human heads from \numIDs different identities.
    \item We propose a new neural-field-based parametric head representation, which facilitates high-fidelity local details through an ensemble of local implicit models.
    \item We demonstrate that our neural parametric head model can be robustly fit to range data, regularize out noise, and outperform existing models.%
\end{itemize}

\section{Related Work}
\noindent{\bf{3D morphable face and head models.}}
The seminal work of Blanz and Vetter~\cite{blanz1999morphable} was one of the first to introduce a model-based approach to represent variations in human faces using PCA. 
Since the scans were captured in constrained environments, the expressiveness of the model was relatively limited. 
As such, improvements in the registration~\cite{paysan20093d}, as well as the use of data captured in the wild~\cite{booth20173d,booth20163d,ploumpis2019combining}, led to significant advances. 
Thereafter, more advanced face models were introduced, including multilinear models of identity and expression~\cite{bolkart2015groupwise,brunton2014multilinear}, as well as models that combined linear shape spaces with articulated head parts~\cite{FLAME}, and localized approaches~\cite{neumann_sparse}.

With the advent of deep learning, various works focused on extending face and head 3DMMs beyond linear spaces. 
To this end, convolutional neural network based architectures have been proposed to both regress the model parameters and reconstruct the face~\cite{tran2018nonlinear,tran2019learning,tran2019towards, FaceScape, FaceVerse, Li_2020_CVPR}.
At the same time, graph convolutions~\cite{bouritsas2019neural,gong2019spiralnet++}
and attention modules~\cite{chen2021learning} have been proposed to model the head mesh geometry.

\noindent{\bf{Neural field representations.}}
Neural field-based networks have emerged as an efficient way to implicitly represent 3D scenes. 
In contrast to explicit representations (e.g., meshes or voxel grids), neural fields are well-suited to represent geometries of arbitrary topology. 
Park et al.~\cite{park2019deepsdf} proposed to represent a class-specific SDF with an MLP that is conditioned on a latent variable. 
Similarly, Mescheder et al.~\cite{mescheder2019occupancy} implicitly define a surface as the decision boundary of a binary classifier and Mildenhall et al.~\cite{mildenhall2021nerf} represent a radiance field using an MLP by supervising a photometric loss on the rendered images.

\begin{figure*}[htb]
    \centering
    \vspace{-0.3cm}
    \includegraphics[width=\textwidth]{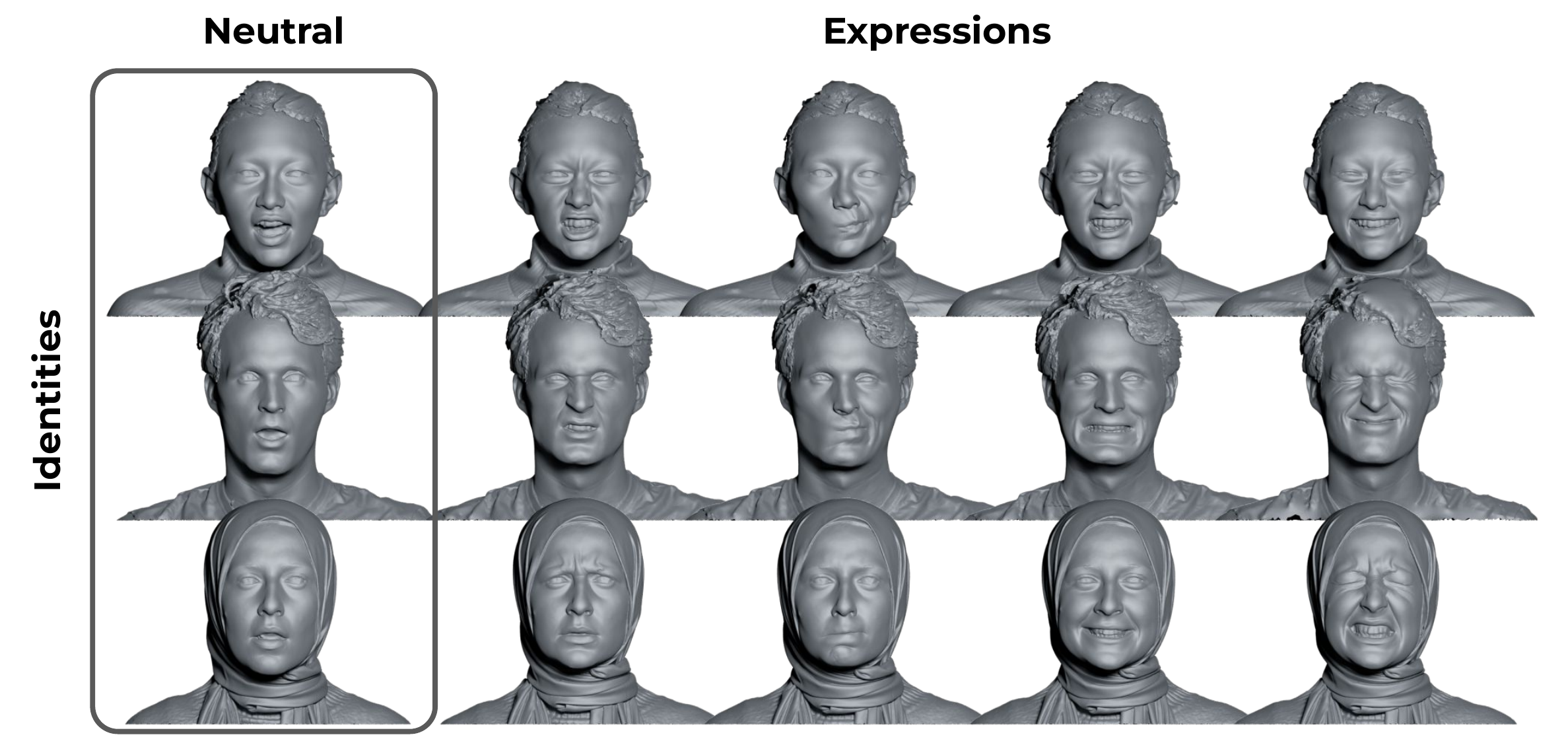}
    \vspace{-0.6cm}
    \caption{3D head scans from our newly-captured dataset: for each person (rows), we first capture a neutral pose, followed by several scans in different expressions (columns). Overall, our dataset has more than \numScans 3D scans from \numIDs people.%
    }
    \label{fig:dataset_figure}
    \vspace{-0.2cm}		
\end{figure*}

Building upon these approaches, a series of works focus on modeling deformations. 
These methods use a separate network to model the deformations that occur in a sequence (e.g.,~\cite{park2021nerfies,hypernerf}), and have been successfully applied to animation of human bodies~\cite{liu2021neural, li2022tava} and heads~\cite{IMavatars}.
Following this paradigm, a number of neural parametric models have been proposed for bodies~\cite{gDNA,NPM,SPAM}, faces~\cite{ImFace}, and ---most closely related to our work--- heads~\cite{i3DMM,wang2022morf,ramon2021h3d}.
For instance, H3D-Net~\cite{ramon2021h3d} and MoRF~\cite{wang2022morf} proposed 3D generative models of heads, but do not account for expression-specific deformations. 
Recently, neural parametric models for human faces~\cite{i3DMM,ImFace} and bodies~\cite{NPM,SPAM,chen2021snarf, gDNA} have explored combinations of SDFs and deformation fields, to produce complex non-linear deformations, while maintaining the flexibility of an implicit geometry representation.
Our work is greatly inspired by these lines of research; however, the key difference is that we tailor our neural field representation specifically to human heads through an ensemble of local MLPs. Thereby, our work is also related to local conditioning methods for neural fields of arbitrary objects \cite{genova2020local, peng2020convolutional, AIR-Nets, DeepLocalShapes}, human bodies \cite{SPAM, zheng2022structured} and faces \cite{ImFace}. Compared to ImFace~\cite{ImFace}, our model %
utilizes a larger number of fine-grained local representations and
incorporates a symmetry prior
to represent the complete head.
Additionally, we propose to models forward instead of backward deformations, which allows for faster animation.

\section{Dataset Acquisition}
\label{sec:data_capture}
Our dataset comprises \numIDs subjects, 29\%  female, and contains over \numScans 3D scans; see Table.~\ref{tab:dataset}.
Our 3D head scans show great level of detail and completeness, as shown in Fig.~\ref{fig:dataset_figure}. 
Additionally, we do not require participants to wear a bathing cap as in the FaceScape dataset~\cite{FaceScape}, allowing for the capture of natural hair styles to a certain degree. See Fig.~\ref{fig:dataset_comparison} for a visual comparison of our novel dataset to other 3D face datasets.
\begin{table}[h]
    \centering
    \begin{tabular}{lc}
			\toprule
			Num. Subjects & 255 (188m/67f) \\
			Total num. Scans    & 5200 \\
			Num. Vertices/Scan & $\approx$ 1.5M\\
			\bottomrule
		\end{tabular}
		
	\vspace{-0.2cm}		
    \caption{Statistics of our 3D scanning dataset.}
    \label{tab:dataset}
    \vspace{-0.2cm}

\end{table}
\subsection{Capture Setup}
Our setup is composed of two Artec Eva scanners~\cite{sivanandan2017assessing}, %
that are rotated 360° around a subject's head using a robotic actuator.
Each scan takes only 6 seconds, which is crucial to keep involuntary, non-rigid facial movements to a minimum. 
The scanners operate at 16 FPS, and are aligned through the scanning sequence and fused into a single mesh; each fused scan contains approximately 1.5M vertices and 3.5M triangles.
Each participant is asked to perform 23 different expressions, which are adopted from the FACS coded expression proposed in FaceWarehouse~\cite{FaceWarehouse}, see our supplemntal for details.
Importantly, we capture a neutral expression with the mouth open, which later serves as canonical expression, as described in Section~\ref{sec:method}.
\newcolumntype{Y}{>{\centering\arraybackslash}X}
\newcolumntype{P}[1]{>{\centering\arraybackslash}p{#1}}
\begin{figure}
    \centering
    \includegraphics[width=\linewidth]{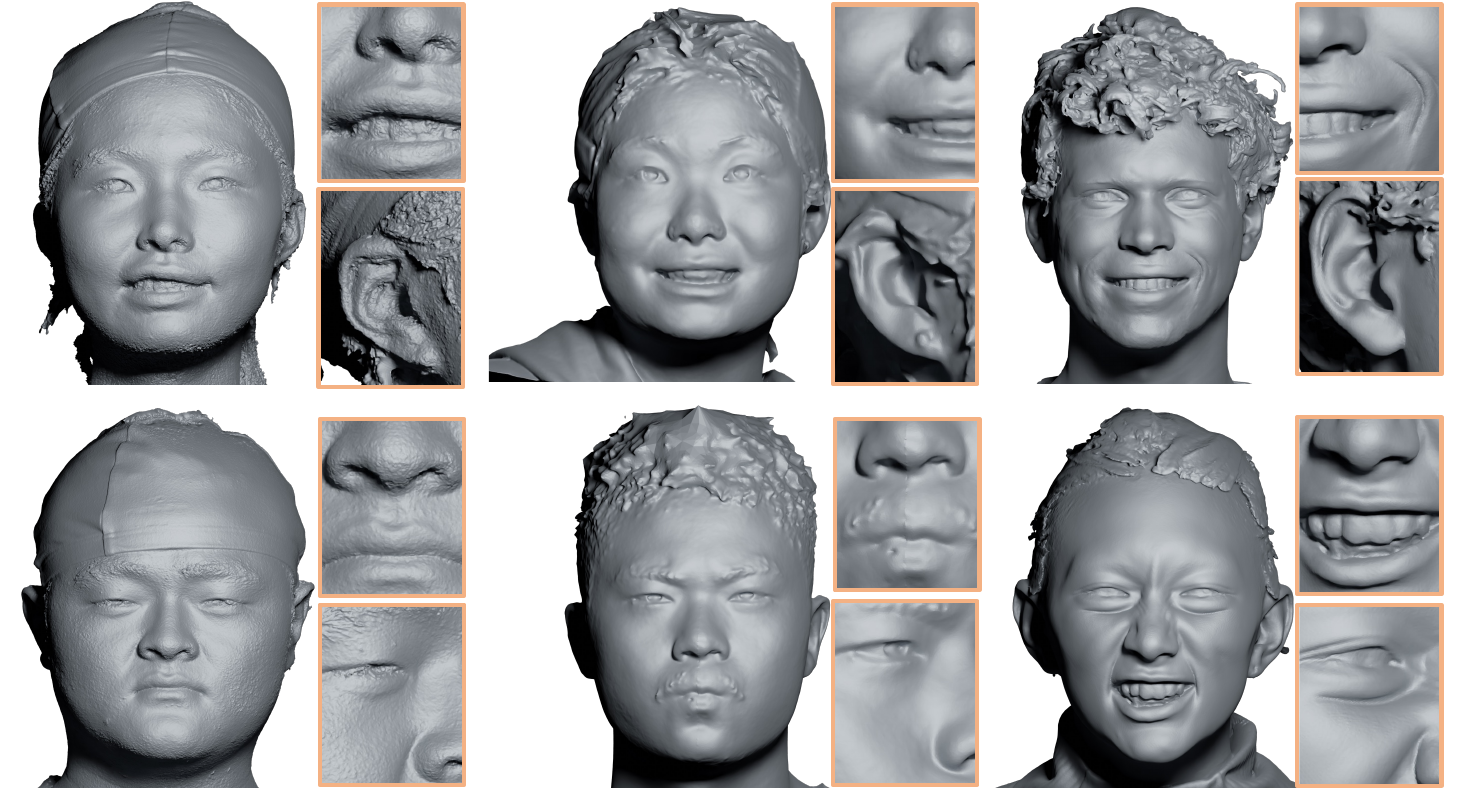}
    \begin{tabularx}{\textwidth}{P{0.25\linewidth}P{0.33\linewidth}P{0.20\linewidth}}
    FaceScape~\cite{FaceScape} & FaceVerse~\cite{FaceVerse} & Ours \\
    \end{tabularx}
    \vspace{-0.4cm}
    \caption{Compared to recent multi-view stero 3D face dataset, our data exhibits sharper details and less noise.}
    \label{fig:dataset_comparison}
\end{figure}
\subsection{Registration Pipeline}
\label{sec:registration}
Registering all head scans against a common template is a key requirement to effectively train our parametric head model. 
First, we start with a rigid alignment into our canonical coordinate system; second, we non-rigidly register all scans to a common template. %
 \begin{figure*}[htb!]
    \centering
    \includegraphics[width=\textwidth]{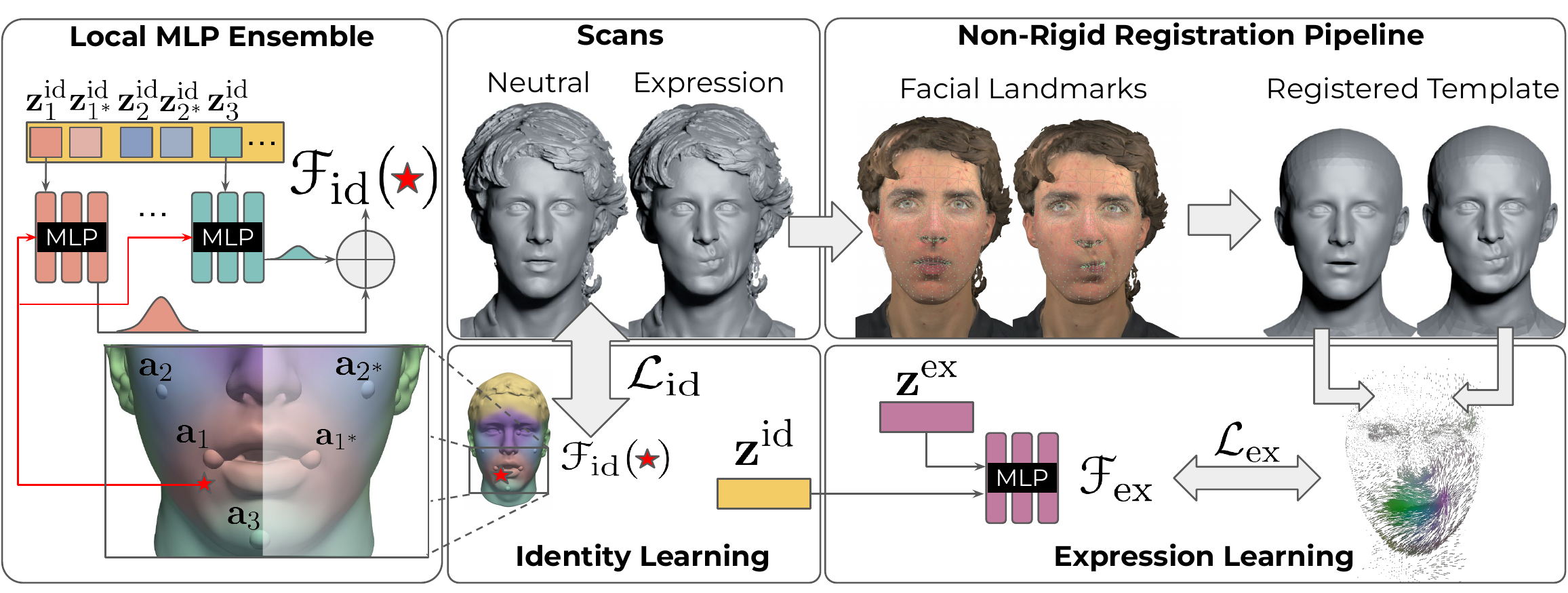}
    \vspace{-0.6cm}
    \caption{Method overview: at the core of our neural parametric head model lies a neural field representation that parameterizes shape and expressions in disentangled latent spaces. Specifically, we propose a local MLP ensemble that is anchored at face keypoints (left).  We train this model by leveraging a set of high-fidelity 3D scans from our newly-captured dataset comprising various expressions for identity (middle). In order to obtain the ground truth deformation samples, we non-rigidly register all scans to a common template (right).}
    \label{fig:pipeline}
\end{figure*}
\subsubsection{Rigid Alignment}
We leverage 2D face landmark detectors to obtain a rigid transformation into the canonical coordinate system of the FLAME model~\cite{FLAME}. 
To this end, we deploy the MediaPipe~\cite{MediaPipe} face mesh detector and back-project a subset of 48 landmarks corresponding to iBUG68 annotations~\cite{sagonas2013300}  to the 3D scan. 
Since not all viewing angles of the scanner's trajectories are suited for 2D facial landmark detection, we instead use frontal renderings of the colored meshes, which yields robust detection quality. 
Note that the initial landmark detection is the only time we use the scanner's color images.
We then calculate a similarity transform using \cite{umeyama} to transform the detected landmarks to the average face of FLAME. %

\subsubsection{Non-Rigid Registration}
As a non-rigid registration prior, we first constrain the non-rigid deformation to FLAME parameter space, before optimizing an offset for each vertex.
Additionally, we back-project 2D hair segmentation masks obtained by FaRL~\cite{FaRL} to mask out the respective areas of the scans.

\paragraph{Initialization.}
Given the 23 expression scans $\{ S_{j}\}_{j=1}^{23}$ of a subject, we jointly estimate identity parameters $\zid \in \mathbb{R}^{100}$, expression parameters $\{ \zex_j\}_{j=1}^{23}$, and jaw poses $\{ \theta_j\}_{j=1}^{23}$ of the FLAME model, as well as a shared scale $s \in \mathbb{R}$ and per-scan rotation and translation corrections $\{R_j\}_{j=1}^{23}$ and $\{t_j\}_{j=1}^{23}$.  
Updating the initial similarity transform is crucial to obtaining a more consistent canonical alignment.

Let $\Phi_j$ denote all parameters affecting the $j$-th FLAME model and $V_{\Phi_j}$ its vertices. 
We jointly optimize for these parameters by minimizing
\begin{equation}
   \argmin_{\Phi_1, \dots \Phi_{23}}\! \sum_{j=1}^{23}\! \left[\lambda_l\Vert L_{j}\! -\! \hat{L}_j \Vert_1\!+\!  d(V_{\Phi_j}, \mathcal{S}_j) \! +\! \mathcal{R}(\Phi_j)\right],
   \label{eq:flame_fitting}
\end{equation}
where $L_j \in \mathbb{R}^{68\times3}$ denotes the back-projected 3D landmarks, $\hat{L}_j$ are the 3D landmarks from $V_{\Phi_j}$, $d(V_{\Phi_j}, S_j)$ is the mean point-to-plane distance from $V_{\Phi_j}$ to its nearest neighbors in scan $S_j$, and $\mathcal{R}(\Phi_j)$ regularizes FLAME parameters.

\paragraph{Fine tuning.}
Once the initial alignment has been obtained, we upsample the mesh resolution by a factor of 16 for the face region, and perform non-rigid registration using ARAP~\cite{ARAP} for each scan individually.

Let $V$ be the upsampled vertices, which we aim to register to the scan $\mathcal{S}$. 
We seek vertex-specific offsets $\{\delta_v\}_{v \in V}$, and auxiliary, vertex-specific rotation $\{R_v\}_{v \in V}$ from the ARAP term. Therefore, we solve
\begin{equation}
     \argmin_{\substack{\{\delta_{v}\}_{v\in V} \\ \{R_v\}_{v \in V}}} \sum_{v \in V}\! \left[ d(\hat{v}, \mathcal{S})\! +\! \sum_{u \in \mathcal{N}_v}\! \Vert R(v\! -\! u)\! -\! (\hat{v}\! - \hat{u})\Vert_2^2 \right],\!
     \label{eq:nrr}
\end{equation}
using the L-BFGS optimizer, where $\hat{v} = v + \delta_v$, $\mathcal{N}_v$ denotes all neighboring vertices, and $d(\hat{v}, \mathcal{S})$ is as before. See the supplemental for more details.

\section{Neural Parametric Head Models}
\label{sec:method}
Our neural parametric head model separately represents geometry in a canonical space and facial expression as forward deformations; see Sections \ref{sec:iden_repr} and \ref{sec:expr_repr}, respectively.

\subsection{Identity Representation}
\label{sec:iden_repr}
We represent a person's identity-specific geometry implicitly in its canonical space as a SDF. 
Compared to template-mesh-based approaches, this offers the necessary flexibility that is required to model a complete head with hair. 
In accordance with related work on human body modeling, \eg \cite{NPM, SPAM, gDNA}, we choose a canonical expression with an open mouth to avoid topological issues. 
While a canonical coordinate system already reduces the dimensionality of the learning problem at hand, we further tailor our neural identity representation to the domain of human heads; as described below.
\subsubsection{Local Decomposition}
Instead of globally conditioning the SDF network on a specific identity, we exploit the structure of the human face to impose two important geometric priors. 
First, we embrace the fixed composition of human faces by decomposing the SDF network into an ensemble of several smaller local MLP-based networks, which are defined around certain facial anchors, as shown in Fig.~\ref{fig:pipeline}. 
Thereby, we reduce the learning problem into smaller, more tractable ones.%
We choose facial anchor points as a trade-off between the relevance of an area and spatial uniformity.
Second, we exploit the symmetry of the face by only learning SDFs on the left side of the face, which are shared with the right half after flipping spatial coordinates accordingly.
More specifically, we divide the face into $K=2K_{\text{symm}} + K_{\text{middle}}$ regions, which are centered at facial anchor points $\mathbf{a} \in \mathbb{R}^{K\times 3}$. 
We use $\mathcal{M}$ to denote the index set anchors lying on the symmetry axis,  and $\mathcal{S}$ and $\mathcal{S}^*$ for symmetric regions on the left and right side respectively, such that for $k  \in \mathcal{S}$ there is a $k^* \in \mathcal{S}^*$ that corresponds to the symmetric anchor point.

In addition to a global latent vector $\zglob \in \mathbb{R}^{d_{\text{glob}}}$, the $k$-th region is equipped with a local latent vector $\zid_k \in \mathbb{R}^{d_{\text{loc}}}$. 
Together, the $k$-th region is represented by a small MLP
\begin{align}
f_k: \mathbb{R}^{d_{\text{glob}} + d_{\text{loc}} + 3} &\rightarrow \mathbb{R} \\ 
   (x, \zidglob, \zid_k) &\mapsto \operatorname{MLP}_{\theta_k}([x - \mathbf{a}_k, \zidglob, \zid_k]),
\end{align}
that predicts SDF values for points $x \in \mathbb{R}^3$, where $[\cdot]$ denotes the concatenation operator. %

In order to exploit face symmetry, we share the network parameters and mirror the coordinates for each pair $(k, k^*)$ of symmetric regions:
\begin{equation}
    f_{k^*}(x, \zidglob, \zid_{k^*}) := f_{k}(\operatorname{flip}(x - a_{k^*}), \zidglob, \zid_{k^*}),
\end{equation}
where $\operatorname{flip}(\cdot)$ represents a flip of the coordinates along the face symmetry axis. 
\subsubsection{Global Blending}
\label{sec:blending}
In order to facilitate a decomposition that helps generalization, it is crucial that reliable anchor positions $\mathbf{a}$ are available. 
To this end, we train a small $\operatorname{MLP}_{\text{pos}}$ that predicts $\mathbf{a}$ from the global latent $\zidglob$. 

Since each local SDF focuses on a specific semantic region of the face, as defined by the anchors $\mathbf{a}$, we additionally introduce $f_0(x, \zidglob, \zid_0) = \operatorname{MLP}_0(x, \zidglob, \zid_0)$, which operates in the global coordinate system, hence covering all SDF values far away from any anchor in $\mathbf{a}$. 
To clarify the notation, we set $a_0 := \mathbf{0} \in \mathbb{R}^3$.

Finally, we blend all local fields $f_k$ into a global field
    \begin{equation}
        \mathscr{F}_{\text{id}}(x) = \sum_{k=0}^{K} w_k(x, a_k) f_k(x, \zidglob, \zid_k),
    \end{equation}
    using Gaussian kernels, similar to \cite{genova2020local, zheng2022structured}, where 
    \begin{equation}
        w_k^*(x, a_k) = 
        \begin{cases}
        e^{\frac{-|| x - a||_2}{2\sigma}}, ~~ \text{if }k > 0\\
                        c, \qquad \qquad\text{if } $k = 0$
        \end{cases}
    \end{equation}
    \begin{equation}
        \text{ and } w_k(x, a_k) = \frac{w_k^*(x, a_k)}{\sum_{k^{\prime}} w_{k^{\prime}}^*(x, a_{k^{\prime}})}
    \end{equation}
    We use a fixed isotropic kernel with standard deviation $\sigma$ and a constant response $c$ for $f_0$.

\subsection{Expression Representation}
\label{sec:expr_repr}

In contrast to our local geometry representation, we model expressions only with a globally conditioned deformation field; \eg a smile will effect the cheeks corners of the mouth and eye region.
In this context, we define $\zex \in \mathbb{R}^{d_{\text{ex}}}$ as a latent expression description. 
Since such a deformation field is defined in the ambient Euclidean space, it is crucial to additionally condition the deformation network with an identity feature.
By imposing an information bottleneck on the latent expression description, the deformation network is then forced to learn a disentangled representation of expressions.

More formally, we model deformations using an MLP
\begin{equation}
    \mathscr{F}_{\text{ex}}(x, \zex, \hat{\z}^{\text{id}}): \mathbb{R}^{d_{\text{ex}} + d_{\text{id-ex}}} \rightarrow \mathbb{R}^{3}.
\end{equation}
Rather than directly feeding all identity information into $\mathscr{F}_{\text{ex}}$ directly, we first project the information to a lower dimensional representation
\begin{equation}
    \hat{\mathbf{Z}}^{\text{id}} = W[\zidglob, \zid_0, \dots \zid_K, \mathbf{a}_1, \dots, \mathbf{a}_K],
\end{equation}
using a single linear layer $W$, where $d_{\text{id-ex}}$ denotes the dimensionality of the interdependence of identity and expression.

\subsection{Training Strategy}
\label{sec:training}

Our training strategy closely follows NPMs~\cite{NPM} and sequentially trains the identity and expression networks in an auto-decoder fashion. 

\medskip
\noindent{\bf{Identity Representation}}
For the identity space, we jointly train latent codes $\mathbf{Z}^{\text{id}}_j := \{\zid_{\text{glob}, j}, \zid_{0, j}, \dots, \zid_{K, j} \}$ for each $j$ in the set of training indices $J$ and network parameters $\theta_{\text{pos}}$ and $\theta_0, \dots, \theta_K$, by minimizing
\begin{equation}
   \mathcal{L}_{\text{id}} = \sum_{j \in J} \mathcal{L}_{\text{IGR}} + \lambda_{a}\Vert \hat{\mathbf{a}}_j - \mathbf{a}_j\Vert_2^2 + \lambda_{\text{sy}}\mathcal{L}_{\text{sy}} + \lambda_{\text{reg}}^{\text{id}}\Vert \mathbf{Z}^{\text{id}}_j \Vert_2^2,
   \label{eq:loss_id}
\end{equation}
where $\mathcal{L}_{\text{IGR}}$ is the loss introduced in \cite{IGR} which enforces SDF values to be zero on the surface and contains an Eikonal term.
This ensures consistency between surface normals and SDF gradients and is in similar spirit to \cite{IGR, SIREN}. 
For training, we directly sample points and surface normals from our ground truth scans.

Additionally, we supervise anchor predictions $\mathbf{a}_j$ using the corresponding vertices from our registrations $\hat{\mathbf{a}}_j$. 
The last two terms serve regularization purposes, where
\begin{equation}
    \mathcal{L}_{\text{sy}} = \sum_{k \in \mathcal{S}}\Vert \zid_k - \zid_{k^*} \Vert_2^2
\end{equation}
enforces the local latent description of symmetric regions to be close, and the final term encourages a well-behaved distribution of both global and local latent descriptions centered around zero.

\medskip
\noindent{\bf{Expression Representation}}
Once the identity representation is learned, we optimize for network parameters $\theta_{\text{ex}}$, $W$ and latent expression codes, \{$\zex_{j, l}\}_{j \in J, l \in L}$, where $j$ indexes identity and $l$ indexes expressions. 
The deformation loss
\begin{equation}
    \mathcal{L}_{\text{ex}}\! =\! \sum_{\substack{i,j \in J,L \\ x \in X_{j, l} }}\! \Vert \mathscr{F}_{\text{ex}}(x, \zex_{j, l}, \hat{\mathbf{z}}_{j}^{\text{id}}) - \delta(x)_{j, l}\Vert_2^2\! +\! \lambda_{\text{reg}}^{ex}\Vert \zex_{j, l} \Vert_2^2
\end{equation}
directly supervises the deformation field using samples $x~\in~X_{j, l}$, which have been precomputed from the registration. See the supplemental for more details. %

\section{Results}
We aim to evaluate how well our method generalizes from our training dataset of 87 identities to unseen ones, and their unique expressions.
Our test dataset consists of 6 female and 12 male identities in 23 expressions each.
We fit our model and baselines to frontal single view depth maps, which are generated by rendering the unseen validation meshes and randomly sampling 5000 points. For ablations with respect to the number of points and noise level, as well as for a demonstration of real-world tracking with NPHM using a commodity depth sensor, we refer to the supplementary material.
In our evaluation, we isolate the reconstruction of identity and expression in section \ref{sec:fitting_id} and \ref{sec:fitting_ex}, respectively.

\medskip
\noindent{\bf Mesh-Based Baselines.}
We evaluate against the Basel Face Model (BFM) and FLAME as representatives of existing template-based PCA-models. Furthermore, we compare against a PCA model with delta expressions~\cite{blanz1999morphable} trained on our registered meshes and a local variant thereof. For the local PCA model we utilize the same facial anchors as in NPHM to divide each neutral registered mesh into regions, which are separately represented by local PCA models. To obtain a final prediction we use the same blending scheme as described in Section \ref{sec:blending}. %
For all these models we additionally provide the 68 facial landmarks as input.%

\medskip
\noindent{\bf Implicit Baselines.}
We compare against ImFace~\cite{ImFace} as a neural backward deformation baseline. To this end, we evaluate a variant of ImFace trained on the FaceScape dataset~\cite{FaceScape} and one that we train on our dataset using their preprocessing (denoted as ImFace*). 
Additionally, we compare against NPMs~\cite{NPM}, isolating the effect of our proposed identity representation.

\medskip
\noindent{\bf Metrics.}
To evaluate the quality of the reconstructions, we report $L_1$-Chamfer distance, normal consistency (N. C.), and F-Score with a threshold of 1.5mm. 

\newcolumntype{Y}{>{\centering\arraybackslash}X}
\newcolumntype{P}[1]{>{\centering\arraybackslash}p{#1}}
\begin{figure*}[htb]
    \centering
    \includegraphics[width=\textwidth]{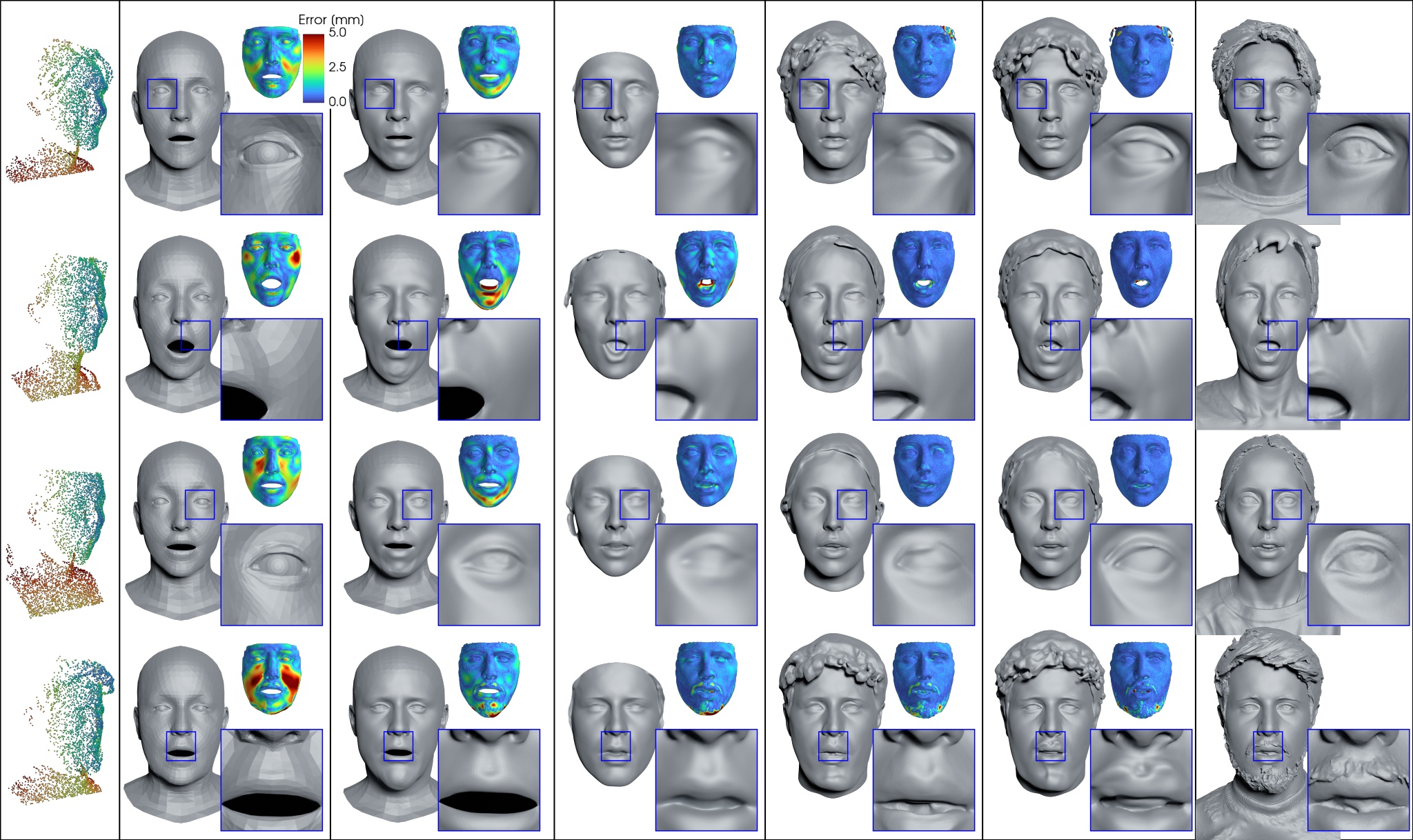}
     \begin{tabularx}{\textwidth}{
         P{0.07\textwidth}%
         p{0.12\textwidth}%
         p{0.13\textwidth}%
         P{0.135\textwidth}%
         P{0.12\textwidth}%
         P{0.11\textwidth}%
         P{0.155\textwidth}%
     }
    Input & FLAME~\cite{FLAME} & Local PCA~\cite{blanz1999morphable} & ImFace*~\cite{ImFace} & NPM~\cite{NPM} & Ours & GT Scan\\
    \end{tabularx}
    \vspace{-0.4cm}
    \caption{Model fitting: at inference time, we fit our model to sparse, partial input point clouds from single depth map. We compare our method to widely-used state-of-the-art parametric face models, including FLAME~\cite{FLAME}, a local PCA~\cite{blanz1999morphable}, ImFace~\cite{ImFace} and neural parametric models (NPM)~\cite{NPM}. Our parametric model has significantly more surface detail and covers the entire head, including the hair region.}
    \label{fig:results_comparison_identity}
\end{figure*}

\subsection{Identity Reconstruction}
\label{sec:fitting_id}
To separately evaluate the quality of our identity space, we fit against a single neutral expression scan for each identity. These scans are aligned to each method's canonical coordinate system. 
We assist baselines that use a closed mouth in their canonical space, i.e., baselines not trained on our data, by optimizing these over all scans instead.
More details on the optimization strategy for each model can be found in the supplemental.

Figure~\ref{fig:results_comparison_identity} and Table~\ref{tab:results_identity} present qualitative and quantitative results, respectively. 
We observe that all neural field methods consistently achieve more faithful reconstructions and further note that the proposed local conditioning allows NPHM to reconstruct details and statistically unlikely elements more reliably.

\begin{table}[h]
    \centering
    \setlength{\tabcolsep}{3pt}
    \begin{tabular}{p{0.27\linewidth}ccc}
			\toprule
			Method&\multicolumn{1}{c}{$L_1$-Chamfer $\downarrow$}&\multicolumn{1}{c}{N. C. $\uparrow$}&\multicolumn{1}{c}{F-Score@1.5 $\uparrow$}\\
			\midrule
			BFM~\cite{paysan20093d}&  $1.341\mathrm{e}{-2}$  &$0.936$&$0.319$\\
			FLAME~\cite{FLAME}&$0.640\mathrm{e}{-2}$  &$0.931$&$0.530$\\

            Global PCA~\cite{blanz1999morphable}&$0.563\mathrm{e}{-2}$  &$0.954$&$0.571$\\
			Local PCA~\cite{blanz1999morphable}&$0.416\mathrm{e}{-2}$  &$0.960$&$0.756$\\

            ImFace~\cite{ImFace}&$0,404\mathrm{e}{-2}$  &$0.954$&$0.832$\\
			ImFace$^*$~\cite{ImFace}&$0.312\mathrm{e}{-2}$  &$0.971$&$0.883$\\

            NPM~\cite{NPM}&  $0.200\mathrm{e}{-2}$  &$0.975$&$0.947$\\
			Ours& $\textbf{0.182}\mathrm{e}{-2}$  &$\mathbf{0.978}$&$\mathbf{0.954}$\\

			\bottomrule
			\multicolumn{2}{l}{\small* trained on our data}\\
		\end{tabular}
		
	\vspace{-0.2cm}	
    \caption{Identity fitting to a single depth map in neutral expression.}
    \label{tab:results_identity}

\end{table}

\subsection{Expression Reconstruction}
\label{sec:fitting_ex}
\begin{figure*}
    \centering
    \vspace{-0.3cm}
    \includegraphics[width=\textwidth]{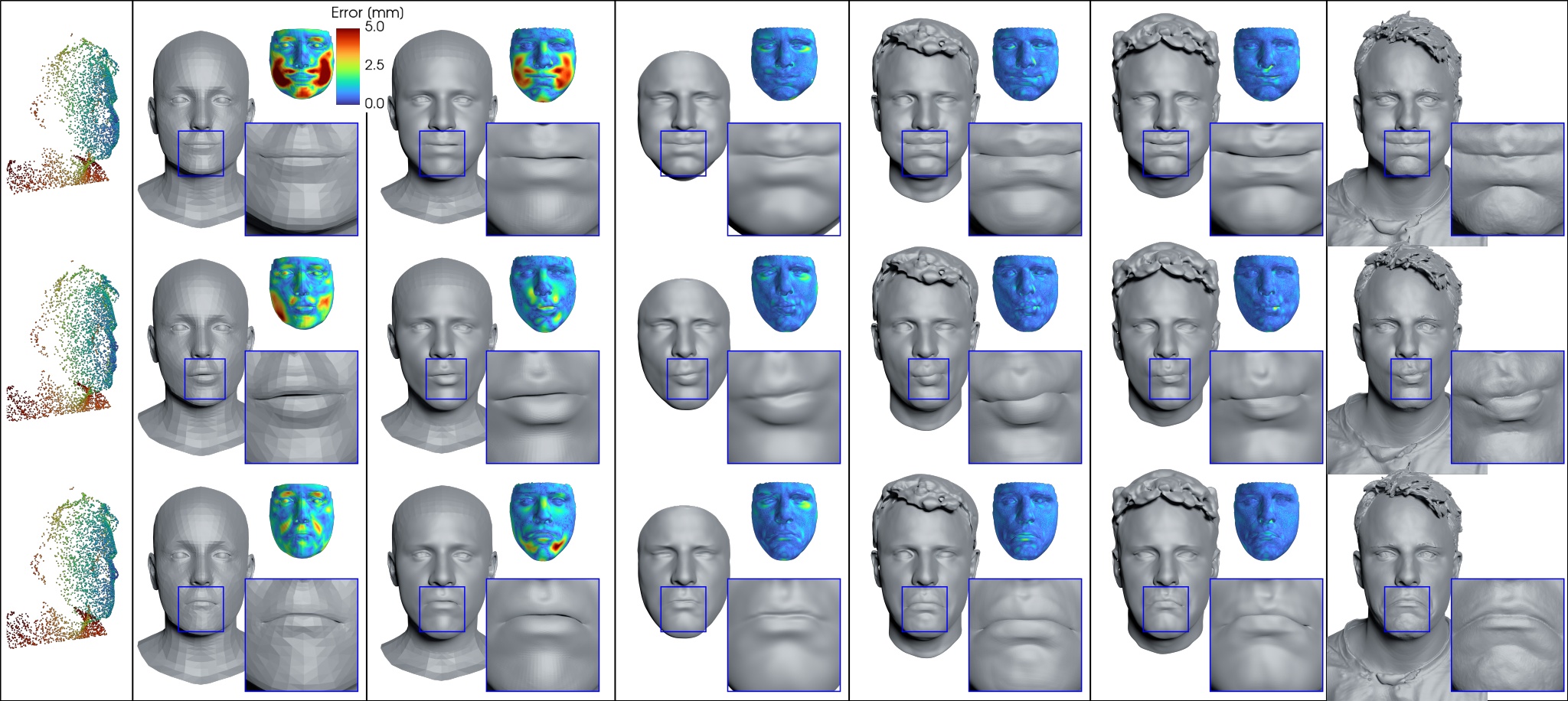}
    \begin{tabularx}{\textwidth}{
        P{0.07\textwidth}%
        p{0.12\textwidth}%
        p{0.13\textwidth}%
        P{0.135\textwidth}%
        P{0.12\textwidth}%
        P{0.11\textwidth}%
        P{0.155\textwidth}%
    }
    Input & FLAME~\cite{FLAME} & Local PCA~\cite{blanz1999morphable} & ImFace*~\cite{ImFace} & NPM~\cite{NPM} & Ours & GT Scan\\
    \end{tabularx}
    \vspace{-0.4cm}
    \caption{Comparison on fitting expressions to sparse input point clouds: from a sparse set of depth observations of different expressions from a frontal view (left), we compare FLAME~\cite{FLAME}, a local PCA~\cite{blanz1999morphable}, ImFace~\cite{ImFace}, neural parametric models (NPM)~\cite{NPM}, and our method against the respective ground truth scans. %
    }
    \label{fig:results_comparison_expression}
\end{figure*}
To evaluate each model's expression space, we fit it to multiple expressions of the same person with the task of recovering one identity code per subject and one expression code per expression. 
For the neural forward deformation models, NPM and NPHM, we utilize iterative root finding \cite{chen2021snarf} to fit the expression codes. For simplicity, we keep the identity code fixed after fitting to the neutral scan. For all other models we jointly solve for expression and identity codes. 
Figure~\ref{fig:results_comparison_expression} and Table~\ref{tab:results_expression} show qualitative and quantitative comparisons with our baselines, respectively.
Owing to the abiility of backward deformations to directly connect the observed with the canonical space, ImFace reliably reconstructs expressions. Nevertheless, it still suffers from blurry reconstructions, compared to both NPM and NPHM. 

See our supplemental for more details and an additional comparison of jointly fitting identity and expression when only a single depth observation is available.

\begin{table}[h]
    \centering
    \setlength{\tabcolsep}{3pt}
    \begin{tabularx}{\linewidth}{p{0.27\linewidth}ccc}
			\toprule
			Method&\multicolumn{1}{c}{$L_1$-Chamfer $\downarrow$}&\multicolumn{1}{c}{N. C. $\uparrow$}&\multicolumn{1}{c}{F-Score@1.5 $\uparrow$}\\
			\midrule
			BFM~\cite{paysan20093d}&     $1.271\mathrm{e}{-2}$&$0.937$&$0.508$\\
			FLAME~\cite{FLAME}& $0.679\mathrm{e}{-2}$&$0.924$&$0.351$\\

   			Global PCA~\cite{blanz1999morphable}& $0.515\mathrm{e}{-2}$&$0.956$&$0.606$\\

			Local PCA~\cite{blanz1999morphable}& $0.535\mathrm{e}{-2}$&$0.950$&$0.641$\\

			ImFace~\cite{ImFace}& $0.369\mathrm{e}{-2}$&$0.959$&$0.824$\\

			ImFace$^*$~\cite{ImFace}& $0.321\mathrm{e}{-2}$&$\mathbf{0.971}$&$0.879$\\
            NPM\cite{NPM}&      $0.299\mathrm{e}{-2}$&$0.962$&$0.891$\\
			Ours&               $\mathbf{0.272}\mathrm{e}{-2}$&$0.969$&$\mathbf{0.913}$\\
			\bottomrule
			\multicolumn{2}{l}{\small* trained on our data}\\
		\end{tabularx}
		
	\vspace{-0.2cm}	
    \caption{Expression fitting on 23 single depth maps per person.}
    \label{tab:results_expression}
    \vspace{-0.2cm}

\end{table}

\subsection{Ablations}
We ablate two main contributions of the proposed identity representation, by fitting identity codes to a neutral scan without involving expressions. First, we analyze the effect of the number of regions $K$ of our ensemble, by comparing against NPM~\cite{NPM}, which effectively would be an ensemble of size 1, and against versions with 12 and 26 regions and adjusted number of latent dimensions.
Additionally, we confirm the benefit of sharing weights for symmetric keypoints. %
Table~\ref{tab:results_ablation} shows a quantitative evaluation of these two ablations supporting our design choices.

\begin{table}[h]
    \centering
    \setlength{\tabcolsep}{3pt}
    \begin{tabular}{lccc}
			\toprule
			Method&\multicolumn{1}{c}{$L_1$-Chamfer $\downarrow$}&\multicolumn{1}{c}{N. C. $\uparrow$}&\multicolumn{1}{c}{F-Score@1.5 $\uparrow$}\\
			\midrule
			NPM~\cite{NPM}&  $0.254$  &$0.972$&$0.906$\\
			K=12, w/ sy.  & $0.289$  &$0.966$&$0.876$\\
			K=26, w/ sy.   & $0.237$  &$0.971$&$0.913$\\
			K=39, w/o sy. & $0.230$  &$0.974$&$0.917$\\
			Ours & $\textbf{0.206}$  &$\mathbf{0.976}$&$\mathbf{0.938}$\\
			\bottomrule
		\end{tabular}
		
	\vspace{-0.2cm}	
    \caption{Effect of the number of anchor points $K$ and symmetry on identity reconstruction performance. NPM represents the extreme case of using exactly 1 anchor point. Note that to be consistent with the original version, NPM differs to the other models in both width and depth of the underlying MLP.}
    \label{tab:results_ablation}

\end{table}

\subsection{Limitations}

In our experiments, we show that NPHM can reconstruct high-quality human heads; however, at the same time, we believe that there are still several limitations and opportunities for future work.
For instance we focus solely on the geometry of heads while omitting any information about appearance. This makes our model ill-suited for fitting  to RGB images using dense photometric terms.
Here, an interesting future avenue would be to explore learning appearance, anchored on top of the geometric base model.
In fact, as part of our dataset we also provide the RGB frames captured during the 3D scanning process, which should facilitate learning such a texture model.

Another limitation is that currently we do not capture loose hair, which limits general diversity; however, compared to other existing face models such as 3D morphable models, we significantly expand the application domain by covering the entirety of the human head.
In the future, we still would like to cover a broader range of hairstyles.

\section{Conclusion}
We have introduced neural parametric head models, a neural representation which disentangles identity and expressions of human heads, by representing geometry in canonical space and modelling expressions as forward deformations.
For our identity representation we have proposed and validated a local representation that is tailored towards human head.
To train our model, we introduce a new dataset of over \numScans high-fidelity 3D scans.
Once trained, our model can be fitted to sparse input point clouds, for instance, captured by a commodity range sensor.
Compared to existing methods, such as widely used PCA-based techniques, our model represents significantly more detail while being able to regularize out noise of the underlying point cloud inputs.
Overall, we believe that our method is an important step towards high-fidelity face capture and our newly introduced dataset opens up opportunities to further explore learning priors for neural face models.

\subsubsection*{Acknowledgements}
This work was supported by the ERC Starting Grant Scan2CAD (804724), the German Research Foundation (DFG) Grant ``Making Machine Learning on Static and Dynamic 3D Data Practical'', the German Research Foundation (DFG) Research Unit ``Learning and Simulation in Visual Computing'', and Synthesia. We would like to thank Maximilian Knörl and Tim Walter for the help with scanning, and Angela Dai for the video voice-over.

{\small
\bibliographystyle{ieee_fullname}
\bibliography{egbib}
}
\appendix

\begin{figure*}[h!]
    \centering
    \includegraphics[width=\textwidth]{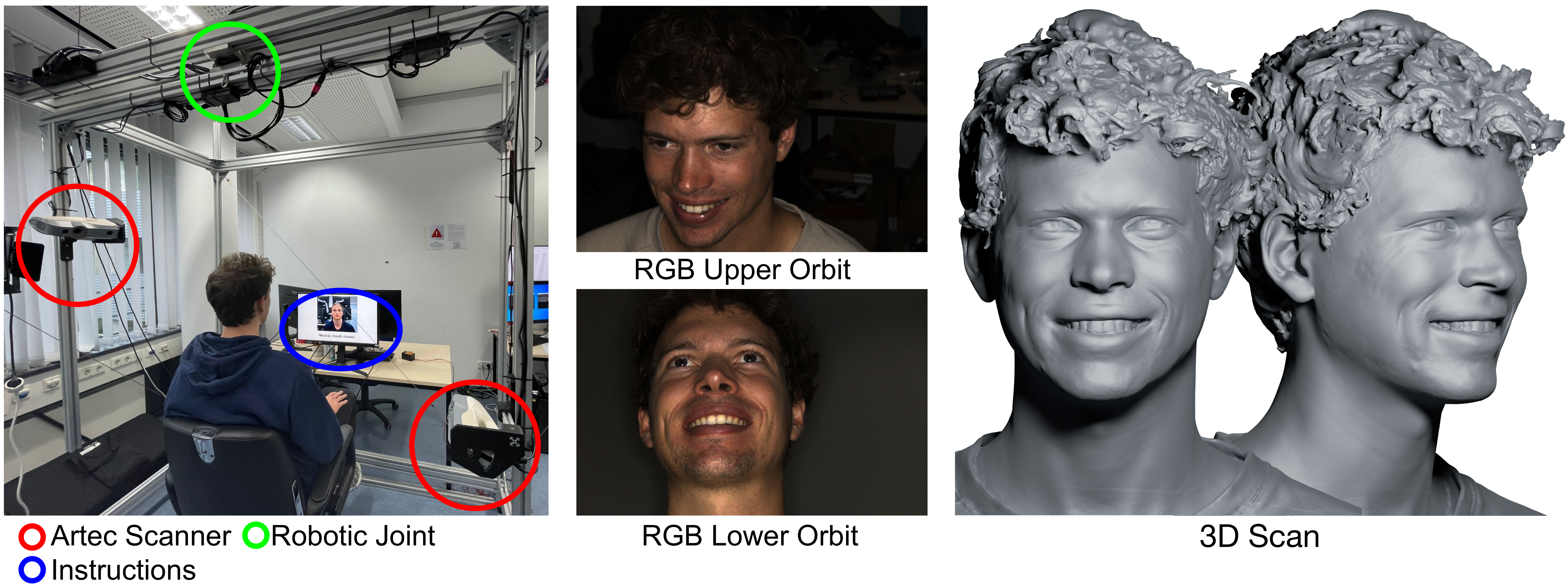}
    \caption{Our custom capture set-up (left). Participants are seated on a height-adjustable chair. A screen presents instructions for the 23 different expression to perform. Next to the resulting 3D scans (right), the scanners also capture 1.3MP RGB images (middle). }
    \label{fig:capture_set_up}
\end{figure*}

\bigskip

\huge \textbf{Appendix}
\normalsize

\section{Overview}

In section \ref{sec:dataset_supp}, we provide additional details about our capture set-up and dataset.
We provide details on the different approaches to fit point clouds for our model and all baselines in Section~\ref{sec:fitting_supp}. Additionally, we provide results on jointly reconstructing an unknown identity and expression from a single point cloud in Section~\ref{sec:fitting_joint_single} and provide a proof-of-concept tracking algorithm for a commodity depth sensor in Section~\ref{sec:tracking}.
Furthermore, we provide implementation details in Section~\ref{sec:details_supp}, and finally we evaluate the robustness of our model with respect to noise and sparsity in Section \ref{sec:ablations_supp}. 
For additional visual results, we refer to our supplemental video. All of our code and data will be available for research purposes. %

\section{Dataset}
\label{sec:dataset_supp}

\begin{figure*}[htb!]
    \centering
    \includegraphics[width=\textwidth]{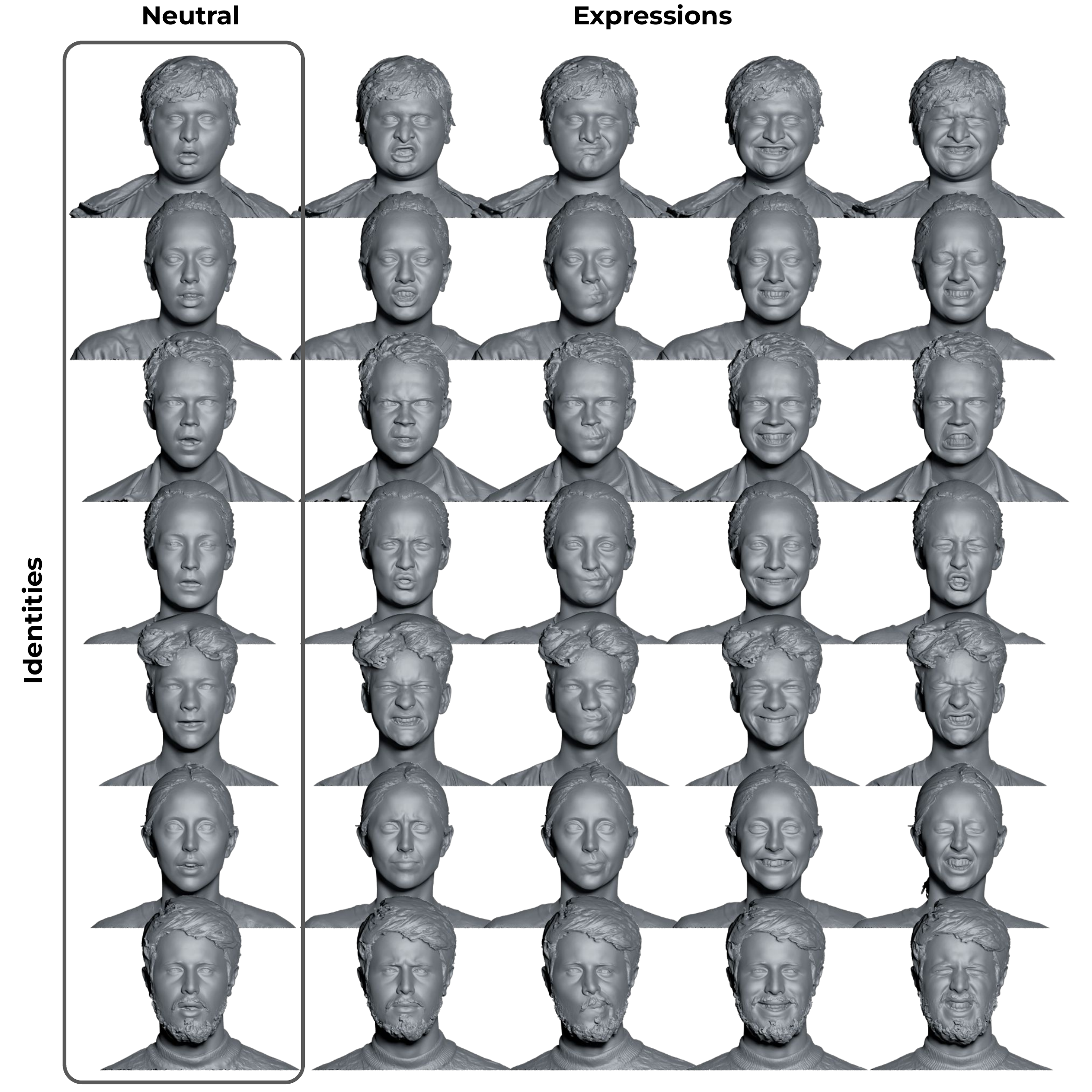}
    \caption{Additional 3D head scans from our newly-captured dataset. Here, we show how different participants perform expressions in their own unique ways. }
    \label{fig:dataset_supp}
\end{figure*}

In order to train our model, we capture a high-quality dataset of 3D head scans. 
To this end, we build a custom scanning setup, which we will detail in the following. 
For samples of our dataset, we refer to Figure~\ref{fig:dataset_supp}.

\subsection{Capture Setup}
Figure~\ref{fig:capture_set_up} shows our custom capture setup, which is built inside of an aluminium cube with an edge length of two meters. We use a robotic actuator\footnote{We use an actuator of the TUAKA series of Sumitomot Drive Technologies: \url{https://us.sumitomodrive.com/en-us/actuators}} to rotate an inverted U-shape around a participant's head. 

We place two Artec Eva scanners opposite of each other, with complementary viewing angles on the ends of the inverted U-shape. The height and angles of the scanners are adjusted to obtain an optimal coverage, while avoiding extreme step angles which decrease scanning accuracy.

\subsection{Details}
During the six seconds of a 360° rotation, each scanner roughly produces 95 frames. For each frame the Artec scanners capture range measurements obtained by analyzing a structured light projection using a stereo camera pair.
Additionally, a third camera captures RGB images every fifth frame, as depicted in Figure \ref{fig:capture_set_up}. Note that we currently do not use the captured RGB input, except for facial landmark detection. 

We process the individual 3D measurements of each frame using the provided software of Artec. First, we align the individual frames of the upper and lower scanner using a global registration algorithm. The individual frames are then fused into a single 3D mesh. Second, we use a hole-filling algorithm and remove disconnected parts.

\subsection{Expressions}
\begin{figure*}[htb!]
    \centering
    \vspace{-0.5cm}
    \includegraphics[width=0.9\textwidth]{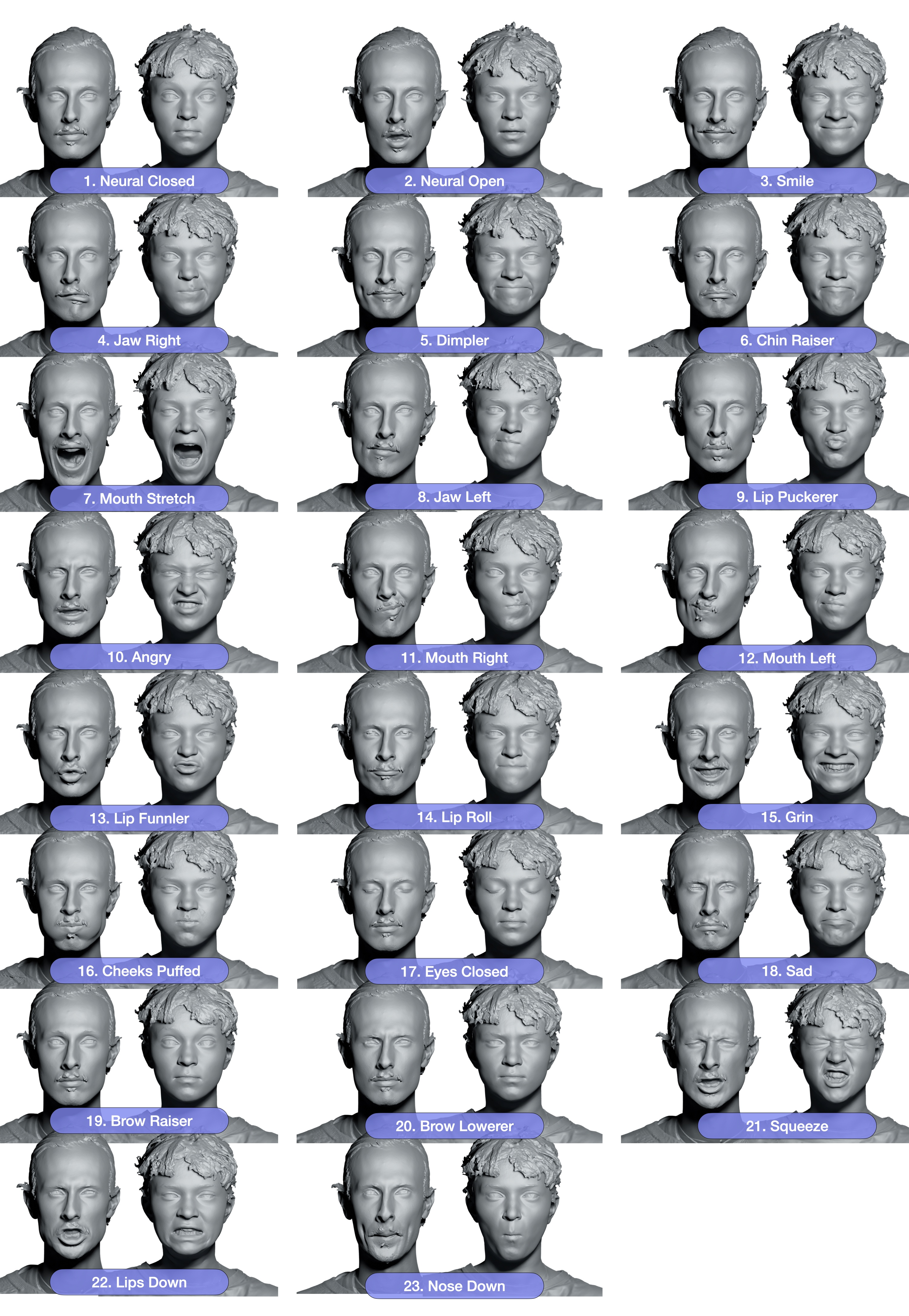}
    \caption{We capture 20 expressions for each participant, and included three bonus expressions for the latest 50 participants. Here, we show two subjects performing all expressions.}
    \label{fig:dataset_supp_expressions}
\end{figure*}

As mentioned in the main paper, our 23 facial expressions are adapted from FaceWarehouse~\cite{FaceWarehouse}. We illustrate the different expressions that we capture in figure \ref{fig:dataset_supp_expressions}. As mentioned before, the neutral, open-mouthed expression is of special importance since it serves as our canonical expression.

\subsection{GDPR}
All participants in our dataset signed an agreement form compliant with GDPR. Please note that GDPR compliance includes the right for every participant to request the timely deletion of their data, which we will enforce as part of the distribution process of our dataset.

\newcolumntype{P}[1]{>{\centering\arraybackslash}p{#1}}
\begin{figure*}
    \centering
    \includegraphics[width=\textwidth]{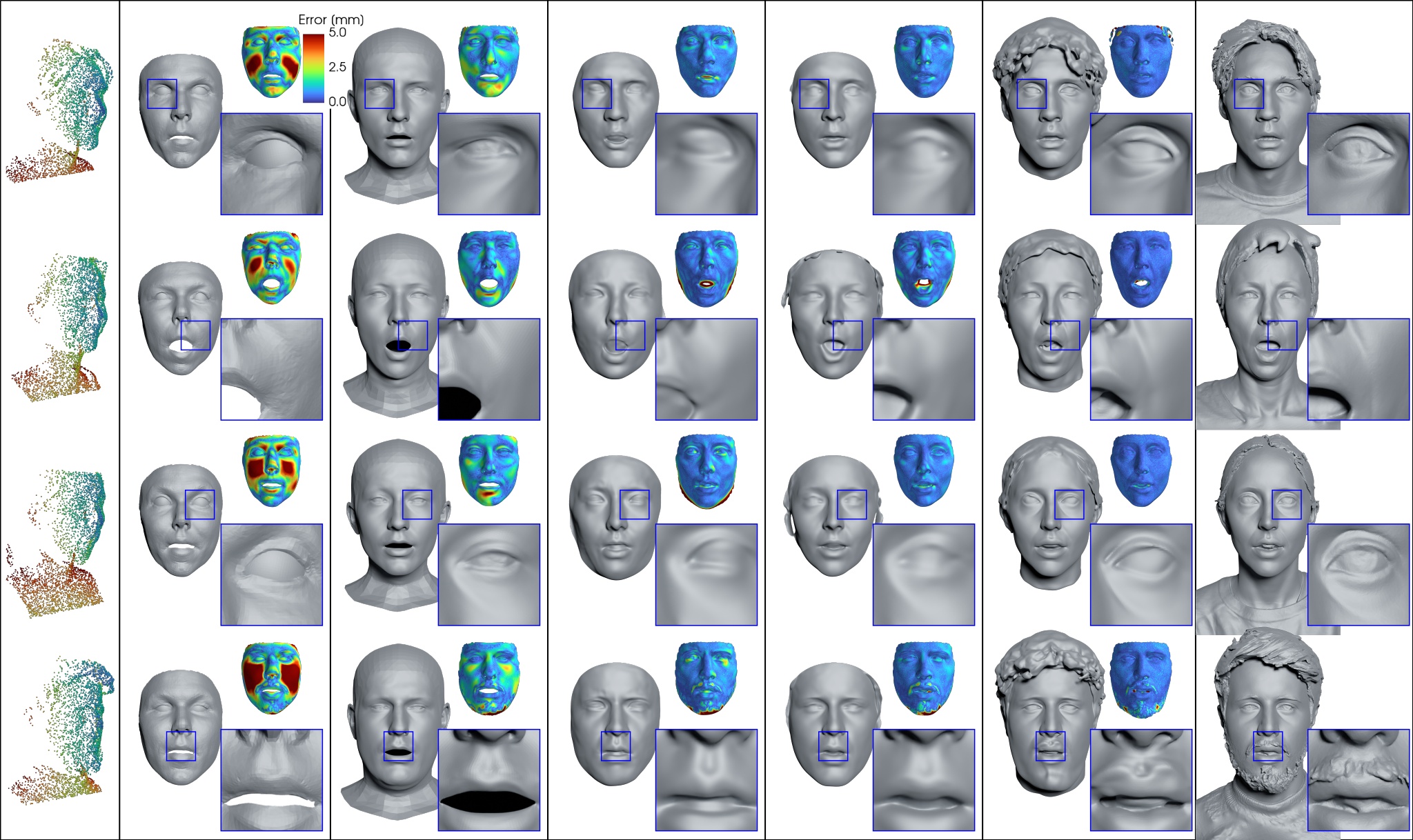}
     \begin{tabularx}{\textwidth}{
        P{0.07\textwidth}%
        P{0.12\textwidth}%
        p{0.13\textwidth}%
        P{0.135\textwidth}%
        P{0.12\textwidth}%
        P{0.11\textwidth}%
        P{0.155\textwidth}%
     }
    Input & BFM~\cite{paysan20093d} & Global PCA~\cite{blanz1999morphable} & ImFace~\cite{ImFace} & ImFace*~\cite{ImFace} & Ours & GT Scan\\
    \end{tabularx}
    \vspace{-0.2cm}
    \caption{More identity fitting comparisons against Basel Face Model~\cite{paysan20093d}, a global PCA~\cite{blanz1999morphable}, ImFace~\cite{ImFace} and an ImFace model that is trained on our data (marked with *). These are the remaining baselines that are missing in Figure \ref{fig:results_comparison_identity} of the main paper.}
    \label{fig:supp_results_comparison_identity}
\end{figure*}
\begin{figure*}
    \centering
    \includegraphics[width=0.99\textwidth]{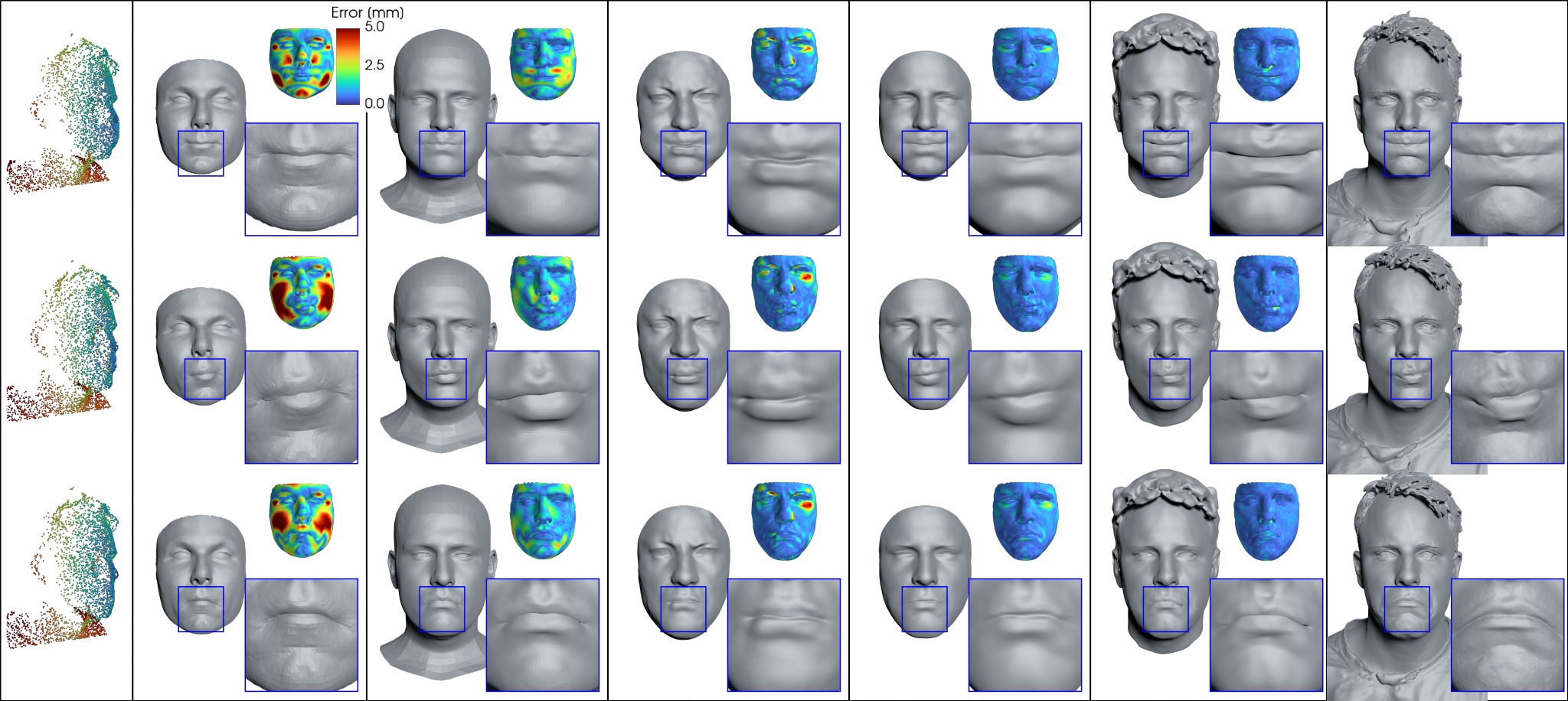}
     \begin{tabularx}{\textwidth}{
        P{0.07\textwidth}%
        P{0.12\textwidth}%
        p{0.13\textwidth}%
        P{0.135\textwidth}%
        P{0.12\textwidth}%
        P{0.11\textwidth}%
        P{0.155\textwidth}%
     }
    Input & BFM~\cite{paysan20093d} & Global PCA~\cite{blanz1999morphable} & ImFace~\cite{ImFace} & ImFace*~\cite{ImFace} & Ours & GT Scan\\
    \end{tabularx}
    \vspace{-0.2cm}
    \caption{More expression fitting comparisons against Basel Face Model~\cite{paysan20093d}, a global PCA~\cite{blanz1999morphable}, ImFace~\cite{ImFace} and an ImFace model that is trained on our data (marked with *). These are the remaining baselines that are missing in Figure \ref{fig:results_comparison_expression} of the main paper.}
    \label{fig:supp_results_comparison_expression}
\end{figure*}

\section{Fitting}
\label{sec:fitting_supp}

In the following, we detail how we use the learned prior of our model and of baselines to fit the models to a single depth frame.
Additionally, we show qualitative results of the remaining baselines for the identity and expression fitting experiment in Figure \ref{fig:supp_results_comparison_identity} and \ref{fig:supp_results_comparison_expression}, respectively.
Furthermore, we present quantitative and qualitative results for joint identity and expression reconstruction based on a single depth map in Section \ref{sec:fitting_joint_single}.

\subsection{Baselines Trained on other Datasets}
Due to the difference in neutral expressions between our model and baselines that were trained on other datasets, \ie BFM~\cite{paysan20093d}, FLAME~\cite{FLAME}, and ImFace~\cite{ImFace}, we cannot fit the identity in an isolated fashion, since that would be unfair. To mitigate this, we fit all these models jointly to all expressions of a person. Additionally, we provide facial landmarks and optimize for Equation \ref{eq:flame_fitting} of the main paper. The results are then used to evaluate both the identity and expression fitting experiments. 
For all other models the fitting procedures are described in the following.

\begin{figure*}[h]
    \centering
    \includegraphics[width=\textwidth]{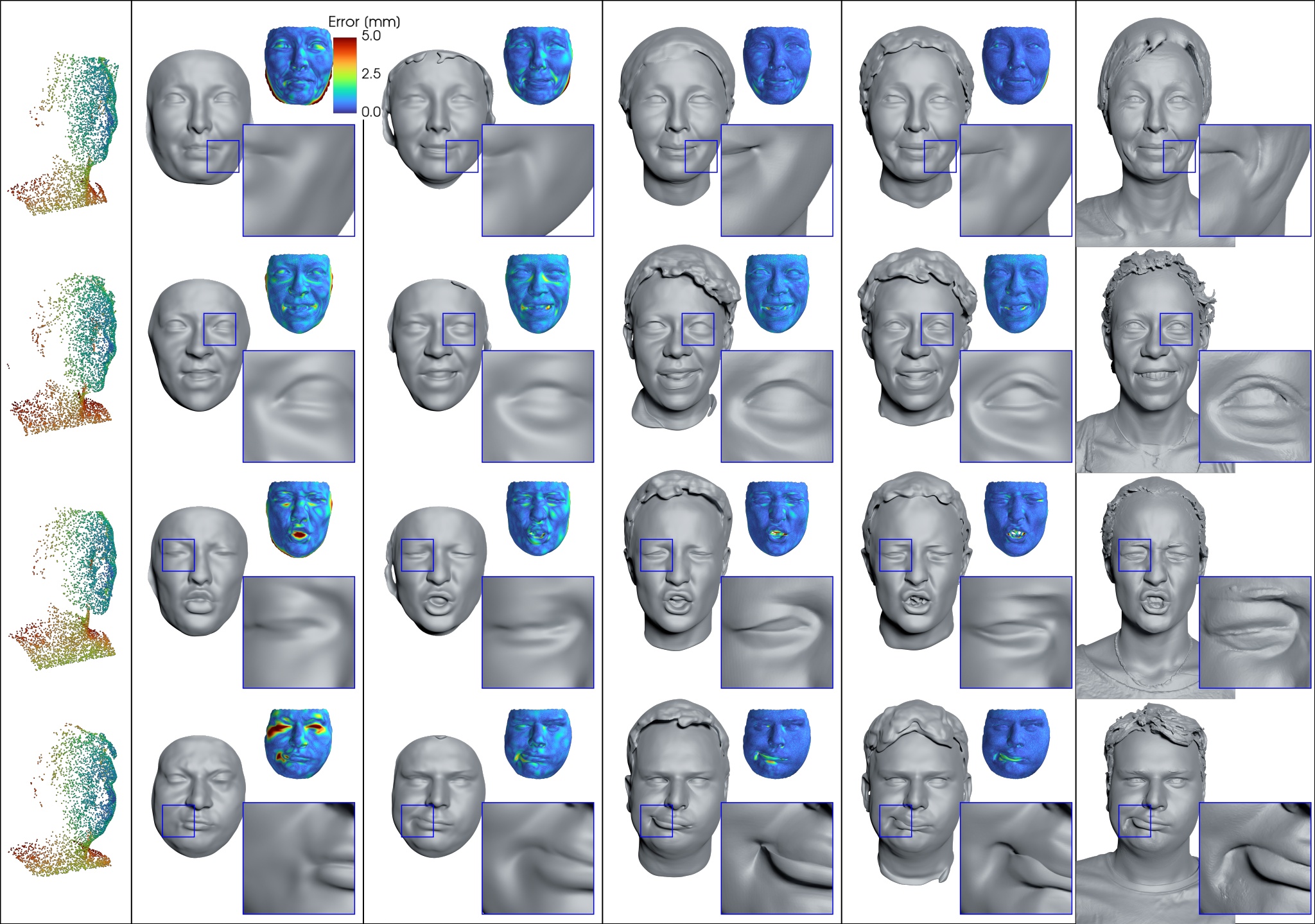}
     \begin{tabularx}{\textwidth}{
        P{0.07\textwidth}%
        P{0.15\textwidth}%
        P{0.17\textwidth}%
        P{0.15\textwidth}%
        P{0.15\textwidth}%
        P{0.155\textwidth}%
     }
    Input & ImFace~\cite{ImFace} & ImFace*~\cite{ImFace} & NPM~\cite{NPM} & Ours & GT Scan\\
    \end{tabularx}
    \vspace{-0.2cm}
    \caption{Results when fitting both identity and expression codes jointly on a single depth map. We compare against ImFace~\cite{ImFace}, an ImFace that is trained on our data (marked with *), and NPM~\cite{NPM}.}
    \label{fig:results_comparison_joint_single}
\end{figure*}

\subsection{Identity Fitting}
Given a single view depth map $X_p \subset \mathbb{R}^3$ of an unknown person in neutral facial expression, we optimize for an identity code $\zid$, as well as an expression code $\zex$. We include the latter in the optimization, in order to account for minor deviations from a perfect canonical facial expression.

\paragraph{PCA-Based Models}
For this purpose, we again optimize for equation \ref{eq:flame_fitting} but only provide the neutral depth map and corresponding landmarks.

\paragraph{ImFace}
Since ImFace utilizes backward deformations, the observed points $X_p$ in posed space can be backward-warped into canonical space, where $\mathscr{F}_{\text{id}}$ can directly act on them. Therefore, the fitting task can be formulated naturally to minimize:
\begin{equation}
    \sum_{x_p \in X} |
    \mathscr{F}_{\text{id}}\left(
    \mathscr{F}_{\text{ex}}^{\leftarrow} (x_p, \zex), 
    \zid
    \right)| + \lambda \mathcal{R}_1(\zid, \zex),
    \label{eq:fitting_backward}
\end{equation}
where $\mathcal{R}_1$ includes the same regularization terms used in ImFace~\cite{ImFace}. We write $\mathscr{F}_{\text{ex}}^{\leftarrow}$ with an arrow to denote backward-direction of the deformation field of ImFace and $\mathscr{F}_{\text{id}}$ for its SDF in canonical space. 
Note that due to simplicity of discussion, we ignore the fact that their $\mathscr{F}_{\text{id}}$ is composed of another deformation field and a template SDF. 
We use the authors' official code and hyperparameters.

\paragraph{NPM and NPHM}
For forward deformation models, formulating a loss to jointly optimize for $\zid$ and $\zex$ is non-trivial. The authors of NPM~\cite{NPM} proposed a formulation that uses a TSDF grid estimated from the depth observations. 
Instead, we resort to the iterative root finding scheme proposed in SNARF~\cite{chen2021snarf}, that inverts the forward deformation.
Given a point $x_p \in X_p$ in posed space, its corresponding points in canonical space is its preimage under the forward-deformation $\mathscr{F}_{\text{ex}}^{\rightarrow}$ . 
The authors of \cite{chen2021snarf} propose to solve for 
\begin{equation}
x_c = \argmin_x | x_p - \mathscr{F}_{\text{ex}}^{\rightarrow}(x, \zex)|
\end{equation}
iteratively to establish a corresponding point $x_c$ in canonical space. 
In order to avoid backpropagation through this iterative procedure, they utilize analytical gradients instead, which can be derived as described in \cite{controlling_neural_levelsets}.
Using these correspondences, we can then resort to the loss in equation \ref{eq:fitting_backward}
\begin{align}
\begin{split}
    &\sum_{x_p \in X} |
    \mathscr{F}_{\text{id}}\left(
    x_c, 
    \zid
  \right)|~ + \\
    &\lambda_{\text{glob}}^{\text{fit}}\Vert 
    \mathbf{z}^{\text{id}}_{\text{glob}} \Vert_2^2\!+\!
    \lambda_{\text{ex}}^{\text{fit}}\Vert \zex \Vert_2^2\!+\! \lambda_{\text{loc}}^{\text{fit}}\sum_{k=1}^K\Vert \mathbf{z}^{\text{id}}_{k} \Vert_2^2\!+\!
    \lambda_{\text{sy}}^{\text{fit}}\mathcal{L}_{\text{sy}},
    \label{eq:fitting_root}
    \end{split}
\end{align}
where $x_c$ replaces the result of the backward deformation.
The second line regularizes all latent codes, as well as the difference between symmetric facial regions. For NPM we simply omit the local latent code and symmetry regularization terms. 
Furthermore, we did not observe topological issues and therefore stick with a single initialization $x_{\text{init}} = x_p$ for the iterative root finding.

For our ablation in Section 5.3 of the main paper, as well as, Section \ref{sec:ablations_supp}, we isolate the expression component completely and replace $x_c$ with $x_p$, assuming that the observed pose is perfectly neutral.

\subsection{Expression Fitting}

In our expression fitting experiment, we investigate the models' performance to obtain $\zid$ and $\{\zex_s\}_{s=1}^S$, given $S$ observed point clouds $\{ X_p^s\}_{s=1}^S$ in posed space, where $S$ is the total number of scans per person.

For our PCA-based baselines, as well as both variants of ImFace, we jointly optimize for the parameters of interest using the same losses as in the previous section.

For the forward deformation models, we find the $\zid$ from the identity fitting already provides a good estimate. For simplicity, we then keep $\zid$ fixed and only optimize for $\{\zex_s\}_{s=1}^S$ using equation \ref{eq:fitting_root}.

\subsection{Single-Expression Fitting}
\label{sec:fitting_joint_single}

The expression fitting task in the main paper attempts to evaluate the expressiveness of each model's expression space by constraining the identity codes to remain the same over all scans of one person. 

Here, we show an additional experiment that aims to reconstruct $\zid$ and $\zex$ jointly given only a single depth map of an unknown person in arbitrary expression. 

Table \ref{tab:results_joint} reports quantitative numbers that further support the effectiveness of the proposed model and Figure \ref{fig:results_comparison_joint_single} shows qualitative results.

\begin{table}[h]
    \centering
    \setlength{\tabcolsep}{3pt}
    \begin{tabularx}{\linewidth}{p{0.27\linewidth}ccc}
			\toprule
			Method&\multicolumn{1}{c}{$L_1$-Chamfer $\downarrow$}&\multicolumn{1}{c}{N. C. $\uparrow$}&\multicolumn{1}{c}{F-Score@1.5 $\uparrow$}\\
			\midrule

			ImFace~\cite{ImFace}& $0.375\mathrm{e}{-2}$&$0.966$&$0.825$\\

			ImFace$^*$~\cite{ImFace}& $0.320\mathrm{e}{-2}$&$0.972$&$0.879$\\

            NPM\cite{NPM}&      $0.243\mathrm{e}{-2}$&$0.969$&$0.928$\\
			Ours&               $\mathbf{0.207}\mathrm{e}{-2}$&$\mathbf{0.974}$&$\mathbf{0.947}$\\
			\bottomrule
			\multicolumn{2}{l}{\small* trained on our data}\\
		\end{tabularx}
		
	\vspace{-0.2cm}	
    \caption{Fitting performance from a single depth map of unknown identity and unknown expression.}
    \label{tab:results_joint}

\end{table}

\subsection{Hyperparameters for NPM and NPHM}
We optimize Equation~\ref{eq:fitting_root} using the Adam optimizer for $700$ iterations. The optimization procedure starts with a learning rate of $0.01$ and is decayed by a factor of $10$ after epochs $200$, $350$, and $500$. For our model we use $\lambda_{\text{glob}}^{\text{fit}} = 0.05$, $\lambda_{\text{loc}}^{\text{fit}} = 0.05$ and $\lambda_{\text{ex}}^{\text{fit}} = 0.003$ to regularize the global and local identity and expression components, respectively. Additionally, we encourage symmetry $\lambda_{\text{sy}}^{\text{fit}} = 1.0$ for the first half of iterations and then set $\lambda_{\text{sy}}^{\text{fit}} = 0.0$. Additionally, we divide $\lambda_{\text{loc}}^{\text{fit}}$ and $\lambda_{\text{glob}}^{\text{fit}}$ by a factor of $5$ at epochs $200$ and $500$, such that the model first learns the coarse facial expression before focusing on the details of the identity. 
For NPM, we use the exact same hyperparameters as for our model. However, the local regularization and symmetry prior have no effect.

\subsection{Real-World Tracking}
\label{sec:tracking}
\begin{figure}[h!]
    \centering
    \vspace{-0.3cm}
    \includegraphics[width=\columnwidth]{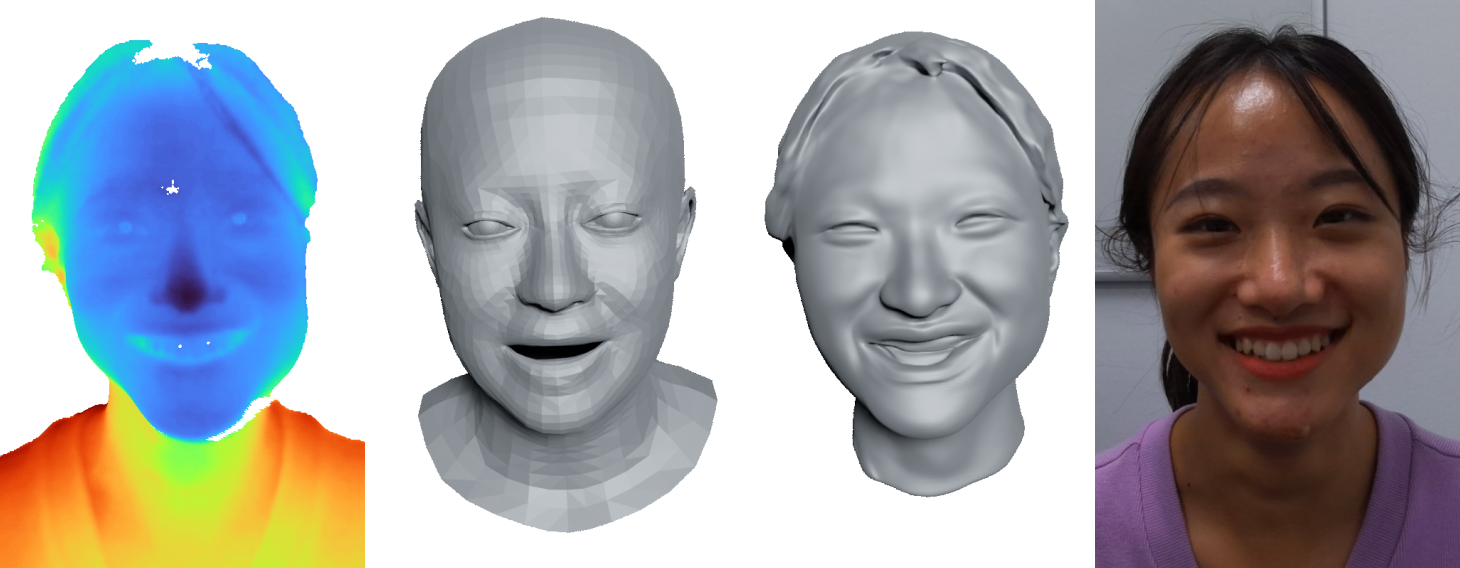}
    \begin{tabularx}{\columnwidth}{P{0.21\columnwidth}P{0.2\columnwidth}P{0.2\columnwidth}P{0.18\columnwidth}}
            Depth Map & FLAME~\cite{FLAME} & Ours & RGB\\
    \end{tabularx}
    \vspace{-0.3cm}
    \caption{Real-world tracking. For a single frame, we show from left to right: the depth map obtained from a commodity depth sensor, FLAME, and our reconstructions, and an image as reference.}
    \label{fig:results_tracking}
    \vspace{-0.3cm}
\end{figure}

Additionally, we evaluate our model in a real-world face tracking scenario.
For this purpose, we fit our model against a depth video captured with a Kinect Azure, a commodity depth sensor. 
Figure \ref{fig:results_tracking} shows our results of a single frame and a comparison to the FLAME model. For the full tracking results, we refer to our supplemental video.

For proof of concept, we optimize for $\zid$ using a single frame and subsequently optimize for head pose and expression parameters for each frame. Additionally, we include a total variation prior along the temporal axis over estimated head pose and expression parameters. More specifically, we add

\begin{equation}
    \mathcal{L}_{TV}(\phi) = \sum_{t=1}^T \Vert \phi(t+1) - \phi(t) \Vert
\end{equation}

to the optimization problem, where $\phi(t)$ denotes any of the time dependent optimization parameters, \ie expression and pose.

To coarsely align the coordinates system of the back-projected depth map into our canonical coordinate system, we calculate the similarity transform using \cite{umeyama} from detected landmarks to the landmarks of the average FLAME face (note that our model shares the  coordinate system of FLAME).

To further guide the optimization, we also include landmarks at the mouth and eye corners, as well as on the top and bottom of the lips, which we denote as $\mathbf{a}_t \in \mathbb{R}^{8 \times 3}$ for each time step. 

First, we utilize the detected landmarks for the initial identity fitting on a chosen frame $t_{\text{can}}$. Here, the landmarks serve as additional supervision for  $\mathbf{z}^{\text{id}}_{\text{glob}}$, by including the term
\begin{equation}
\Vert \text{MLP}_{\text{pos}}(\mathbf{z}^{\text{id}}_{\text{glob}}) - \mathbf{a}_{t_{\text{can}}}\Vert_1.
\end{equation}
In this stage, we additionally estimate normals using a Sobel filter and use them as additional supervision signal; cf.~Equation~\ref{eq:IGR_loss}.

During expression fitting, we incorporated the eight facial landmarks as direct supervision for the forward deformation network:
\begin{equation}
    \sum_{t=1}^T \Vert \mathscr{F_{\text{ex}}}(\text{MLP}_{\text{pos}}(\mathbf{z}^{\text{id}}_{\text{glob}}), \mathbf{z}^{\text{ex}}_t, \mathbf{z}^{\text{id}}_{\text{glob}}) - \mathbf{a}_t \Vert_1.
\end{equation}

\section{Implementation Details}
\label{sec:details_supp}

We implement our approach -- including registration, training, and inference -- in PyTorch and, unless otherwise mentioned, run all heavy computations on the GPU, for which we use an  Nvidia GTX 3090.

\subsection{Non-Rigid Registration}
\label{sec:details_nrr}

In Equations \ref{eq:flame_fitting} 
and \ref{eq:nrr}
of the main paper, we use the point-to-plane distance $d(v, \mathcal{S})$ from a point $v \in \mathbb{R}^3$ to a surface $\mathcal{S} \subset \mathbb{R}^3$. To make our energy terms more robust, we filter this distance based on a distance $\delta_d$ and normal threshold $\delta_n$, such that

\begin{equation}
    d^*(v, \mathcal{S}) = 
    \begin{cases}
        0, \qquad \quad \text{if } d(v, \mathcal{S}) > \delta_d,\\
        0, \qquad \quad \text{if } \langle n(v), n(s) \rangle > \delta_n, \\
        d(v, \mathcal{S}), \quad \text{otherwise},
    \end{cases}
\end{equation}

where 
\begin{equation} 
d(v, \mathcal{S}) = \min_{s\in \mathcal{S}} | \langle v - s, n(s)\rangle |
\end{equation}
is the unfiltered point to plane distance and $n(v)$ and $n(s)$ denote the vertex normals of $v$ in the template mesh and the normals of its nearest neighbor in the target $\mathcal{S}$, respectively.

\paragraph{FLAME Fitting} 
We regularize our optimization in FLAME parameter space using
\begin{align}
    \mathcal{R}(\Phi_j) = \lambda_{\text{id}}\frac{\Vert \mathbf{z}^{\text{id}}\Vert_2^2}{20} &+
                            \lambda_{\text{ex}}\Vert \mathbf{z}^{\text{ex}_j}\Vert_2^2 +  \lambda_{\text{jaw}} \Vert \theta_j\Vert_2^2 \nonumber \\
                            &+ 
                            \lambda_{\text{rigid}} (\Vert R_j\Vert_2^2 + \Vert t_j\Vert_2^2).
\end{align}
We use $\lambda_{\text{id}}= 1/5000$, $\lambda_{\text{ex}} = 1/3000$ to regularize the identity and expression parameters respectively. For the jaw angle and the rigid parameters, we regularize with $\lambda_{\text{jaw}} = 1/10$ and $\lambda_{\text{rigid}} = 1/10$. Since the point-to-plane distance initially gives an unreliable signal, despite our filtering we down-weight the point-to-plane distance with $\lambda_d=1/15$ for the first 300 iterations. For all remaining iterations of the 2000 iterations, we set $\lambda_d=1$. We solve Equation~ \ref{eq:flame_fitting}
using the Adam~\cite{adam} optimizer with a learning rate of $4e^{-3}$, which is decayed by a factor of 5 for the final 500 iterations.

\paragraph{Finetuning}

We exponentially decay the weight $\lambda_{\text{ARAP}}$ of the ARAP~\cite{ARAP} term with a factor of $0.99$. We start with $\lambda_{\text{ARAP}} = 10.0$, but do not decay below $\lambda_{\text{ARAP}} = 0.1$. On average our implementation converges after 400-500 iterations of the L-BFGS optimizer and takes roughly 4 minutes on a single Nvidia 1080 GPU.

Since both the FLAME fitting and finetuning require a large number of nearest neighbor queries between vertices of the optimized mesh and the target mesh, we utilize FAISS~\cite{FAISS}, which provides efficient, GPU-optimized search indices for approximate similarity search.

\subsection{Data Preparation and Training}
\label{sec:details_train_data}

\paragraph{Identity Training}
To train $\mathscr{F}_{\text{id}}$, we use the loss

\begin{align}
\begin{split}
    \label{eq:IGR_loss}
    \mathcal{L}_{\text{IGR}} &= \sum_{x \in \delta X} \lambda_s | \mathscr{F}_{\text{id}}(x)| + \lambda_s\left(1\!-\! \langle \nabla \mathscr{F}_{\text{id}}(x), n(x)\rangle\right)  \\
    &+ \sum_{x \in X \cup \delta X} \lambda_{\text{eik}}(\Vert \nabla \mathscr{F}_{\text{id}(x)}\vert_2\! -\! 1)\\
    &+ \sum_{x \in X}\lambda_{0}\text{exp}(-\alpha |\mathscr{F}_{\text{id}}(x)|)
\end{split}
\end{align}

introduced in \cite{IGR} and \cite{SIREN}, where we omit the conditioning of $\mathscr{F}_{\text{id}}$ for simplicity. Here, $\delta X$ denotes samples on the surface and $X$ denotes samples in space. We choose $\lambda_s = 2$, $\lambda_n = 0.3$, $\lambda_{\text{eik}}= 0.1$ and $\lambda_0=0.01$. For the additional hyperparameters mentioned in Equation~(11)%
we set $\lambda_{\text{reg}}^{\text{id}}=0.005$, $\lambda_a = 7.5$ and $\lambda_{\text{sy}} = 0.005$.

Furthermore, we train for $15,000$ epochs with a learning rate of $0.0005$ and $0.001$ for the network parameters and latent codes, respectively. Both learning rates are decayed by a factor of $0.5$ every $3,000$ epochs. We use a batch size of $16$ and $|\delta X| = 500$ points sampled on the surface. Samples $X$ are obtained by adding Gaussian noise with $\sigma = 0.01$ to surface points and adding some points sampled uniformly in a bounding box. Additionally, we use gradient clipping with a cut-off value of $0.1$ and weight decay with a factor of $0.01$.

Since this loss only requires samples on the surface directly, we precompute $2,000,000$ points sampled uniformly on the surface of the 3D scans, after removing the lower part of the scan, which we determine using a plane spanned by three vertices on the neck of our registered template mesh. Since our focus lies on the front part of the face, 80\% of these points are sampled on the front and 20\% on the back and neck. The frontal area is determined by a region on our registered meshes, which covers the face, ears, and forehead. We additionally sample surface normals.

Training the identity network takes about 12 hours until convergence on a single GPU.

\paragraph{Expression Training}
For the training of  $\mathscr{F}_{\text{ex}}$, we follow NPMs~\cite{NPM} and precompute samples of the deformation field, which can be used for direct supervision of  $\mathscr{F}_{\text{ex}}$.

More specifically, let $\mathcal{M}$ and $\mathcal{M}^{\prime}$ be a neutral and expression scan. For a point $x \in \mathcal{M}$, we determine the corresponding point $x^{\prime} \in \mathcal{M}^{\prime}$ using barycentric coordinates and construct samples of the deformation field $\delta(x) = x^{\prime} - x$. While strictly speaking the deformation is only defined for points on the surface, we compute field values close to the surface by offsetting along the normal direction, \ie $\delta(x + \alpha n(x)) = x^{\prime} + \alpha n(x^{\prime}) - (x + \alpha n(x)) $, where we sample $\alpha \sim \mathcal{N}(0, \tau_{i}\mathbb{I}_3)$ twice with standard deviations $\tau_1 = 0.02$ and $\tau_2 = 0.004$. Overall, we sample $2,000,000$ points per expression.

For the expression training we use $\lambda_{\text{reg}}^{\text{ex}} = 5e^{-5}$ and a learning rate of $5e^{-4}$ and $1e^{-3}$ for the network and latent codes, respectively. We train for $2,000$ epochs with a learning rate decay of $0.5$ every $600$ epochs, gradient clipping at $0.025$ and weight decay strength $5e^{-4}$. We use $1000$ samples to compute $\mathcal{L}_{\text{ex}}$ and a batch size of 32.

Training the expression network until convergence takes about 8 hours on a single GPU.

\begin{figure*}[htb!]
     \centering
     \hfill
     \begin{subfigure}[b]{0.25\textwidth}
         \centering
         \includegraphics[width=\textwidth]{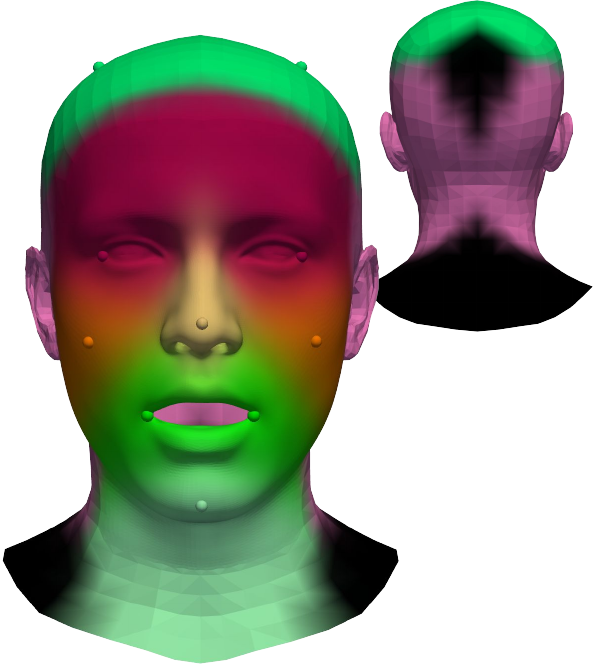}
         \caption{$K=12\qquad$}
         \label{fig:plot_num_points}
     \end{subfigure}
     \hfill
     \begin{subfigure}[b]{0.25\textwidth}
         \centering
         \includegraphics[width=\textwidth]{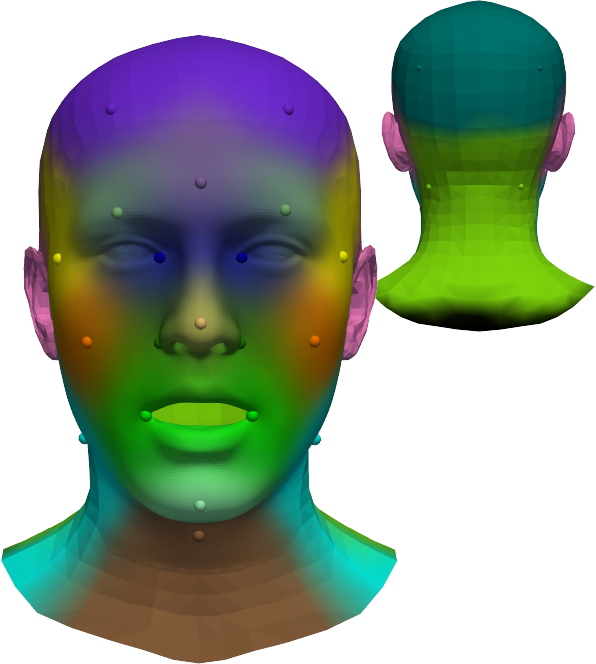}
         \caption{$K=26\qquad$}
         \label{fig:plot_num_points}
     \end{subfigure}
     \hfill
     \begin{subfigure}[b]{0.25\textwidth}
         \centering
        \includegraphics[width=\textwidth]{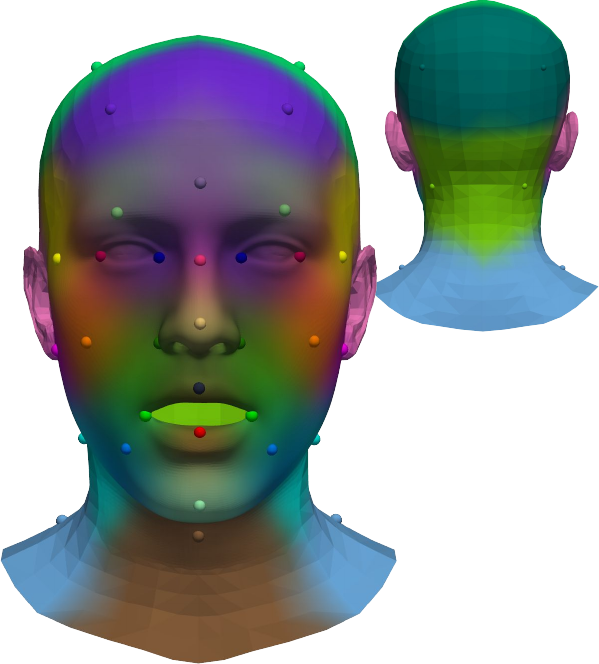}
         \caption{$K=39\qquad$}
         \label{fig:plot_noise}
     \end{subfigure}
     \hfill
     
        \caption{Anchor Layout: Each anchor is assigned a unique color, except for symmetric pairs which share colors. We calculate vertex colors by blending in the same fashion, as for the ensemble of local MLPs. Consequently, the colors show the influence that each local MLP has on its surrounding. Black denotes the color of $f_0$. Anchor points were chosen as vertices of the average over all registrations.}
    \label{fig:anchor_layout}
\end{figure*}

\begin{figure*}[htb!]
     \centering
     \begin{subfigure}[b]{0.49\textwidth}
         \centering
         \includegraphics[width=\textwidth]{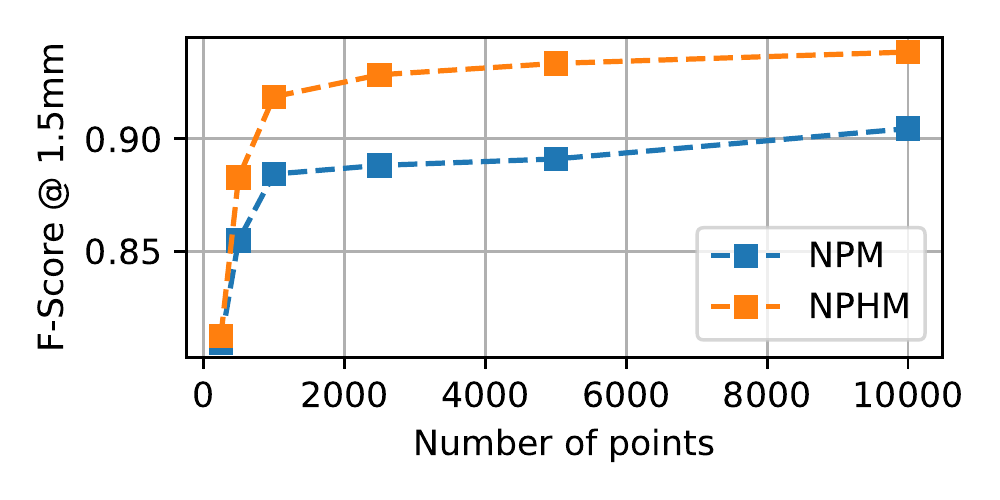}
         \caption{Number of Points}
         \label{fig:plot_num_points}
     \end{subfigure}
     \hfill
     \begin{subfigure}[b]{0.49\textwidth}
         \centering
        \includegraphics[width=\textwidth]{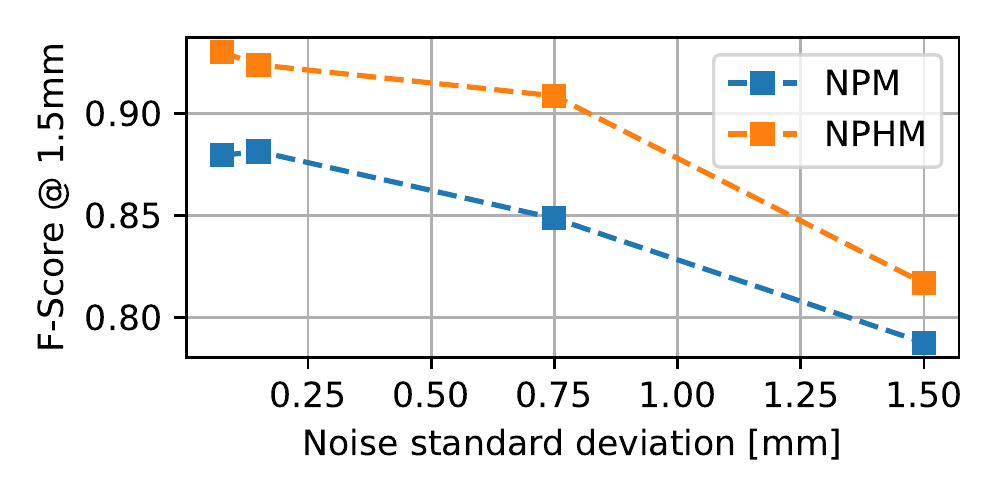}
         \caption{Additive Gaussian Noise}
         \label{fig:plot_noise}
     \end{subfigure}
     
        \caption{Robustness of our method with respect to (a) the number of observed 3D points and (b) additive Gaussian noise to the input point cloud. The results indicate that both NPM and NPHM are similarly effected by worsening quality of observations.}
\end{figure*}

\subsection{Architectural Details}
\label{sec:details_architecture}

\subsubsection{NPMs}
In the main paper, we compare our proposed method against our implementation of NPMs~\cite{NPM}. Instead of the proposed ensemble of local MLPs, NPMs use the original architecture of DeepSDF~\cite{park2019deepsdf} with 8 layers, a hidden dimensionality of 1024, and $\mathbf{Z}_{\text{id}}=512$ dimensions for the latent vector for $\mathscr{F}_{\text{id}}$.

The expression latent dimension is  $d_{\text{ex}} = 200$ and the MLP has 6 hidden layers with 512 hidden units. We use identical settings for NPHM.

\subsubsection{NPHMs}
\label{sec:nphm}
Our default choice for the number of anchor points is $K=39$, of which $K_{\text{symm}}=16$ are symmetric. This leads to $7$ anchor points lying directly on the symmetry axis, and hence parameters of $16+7=23$ local DeepSDFs have to be optimized. Figure~\ref{fig:anchor_layout} depicts the arrangement of the anchor points.

The identity latent space is composed of the shared global part $\zidglob \in \mathbb{R}^{d_{\text{glob}}}$ with $d_{\text{glob}}=64$ and local latent vectors $\zid_{k} \in \mathbb{R}^{d_{\text{loc}}}$ with $d_{\text{loc}}=32$.
Our local MLPs have 4 hidden layers with 200 hidden units each and follow the DeepSDF~\cite{park2019deepsdf} architecture.
Note that the total number of latent identity dimensions $d_{\text{id}} = (K+1)*d_{\text{loc}} + d_{\text{glob}} = 1344$.

Furthermore, we use $\sigma=0.1$ and $c=e^{-0.2/\sigma^2}$ to blend the ensemble of local MLPs. Figure \ref{fig:anchor_layout} illustrates the resulting influence that the individual local MLPs have on the final prediction.

\paragraph{Anchor Points}
In the main paper, we ablated the number of face anchor points. Figure~\ref{fig:anchor_layout} shows a comparison of the different anchor layouts that we ablated. For a lower number of anchors, we increase $d_{\text{loc}}$ such that $d_{\text{id}}$ is roughly preserved.

For the ablation of our symmetry prior, we keep the exact same anchor layout; however, do not share network weights for symmetric anchors, do not mirror coordinates, and do not include the symmetry regularizer during fitting.

\subsection{Metrics}
\label{sec:eval}

Since we quantitatively compare models that represent vastly different regions of the human head, we restrict the calculations of our metrics to the face region.
This also aligns with the fact, that each model only observes a single, frontal depth map, i.e. other parts of the head can only be estimated roughly.

To this end, we determine the facial area by all points which are closer than 1cm to a region defined on our registered template mesh. Within this region, we sample 1,000,000 points with their corresponding normals on the ground truth as well as on each reconstruction. Using these sampled points and normals, we compute all of our metrics. %

Please note, that this evaluation does not account for the fact that reconstructions of closed-mouth expressions might contain the inner part of the mouth. The inner part of the mouth is not visible by the 3D Scanners and hence is missing in the ground truth. This especially is a disadvantage for forward deformation models, since they reconstruct large parts of the inner mouth region. To account for this one might have to exclude sampled points in the reconstructions that are not visible, \eg by rendering depth images from multiple views and backprojecting them to 3D.

\section{Additional Ablations}
\label{sec:ablations_supp}
The experiments in the main paper were restricted to single view depth maps with 5000 points. 
Here, we present a thorough evaluation with respect to the number of input points and with respect to artificial Gaussian noise. Note that these experiments aim to ablate the different identity representations between NPM and NPHM. Henc,  we only perform identity fitting in the following.

\paragraph{Number of Points:} 
Figure \ref{fig:plot_num_points} shows how the number of observed points effect the reconstructions quantitatively. We evaluate on 250, 500, 1000, 2500, 5000, and 10000 points, respectively. Figure~\ref{fig:qual_num_points} illustrates the effect qualitatively.

\paragraph{Noise:}
Similarly, we ablate against additive Gaussian noise with standard deviations of 0.0mm, 0.3mm, 0.75mm and 1.5mm. Quantitative and qualitative results are presented in Figures \ref{fig:plot_noise} and \ref{fig:qual_noise}, respectively.

\newcolumntype{Y}{>{\centering\arraybackslash}X}
\newcolumntype{P}[1]{>{\centering\arraybackslash}p{#1}}
\begin{figure}[tbh!]
    \centering
    \includegraphics[width=\linewidth]{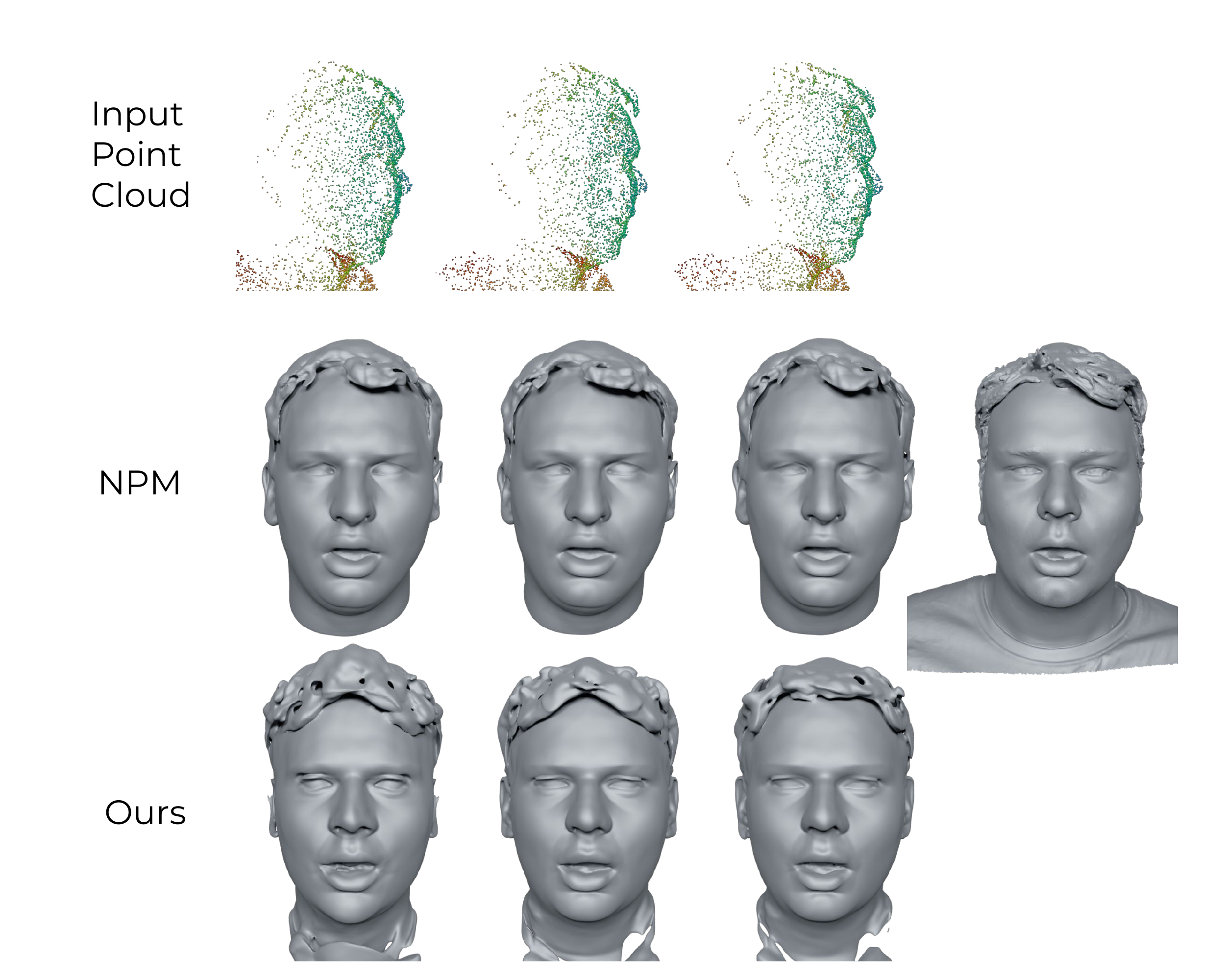}
    \begin{tabularx}{\linewidth}{P{0.14\linewidth}p{0.15\linewidth}P{0.15\linewidth}P{0.12\linewidth}P{0.18\linewidth}}
    Noise Level& 0.75 mm & 0.30 mm & 0 mm & GT Scan\\
    \end{tabularx}
    \caption{Qualitative comparison of NPMs~\cite{NPM} and our method with respect to noise in the input point cloud. We perturb the points by applying random Gaussian noise with different standard deviations. }
    \label{fig:qual_noise}
\end{figure}

\begin{figure}[tbh!]
    \centering
    \includegraphics[width=\linewidth]{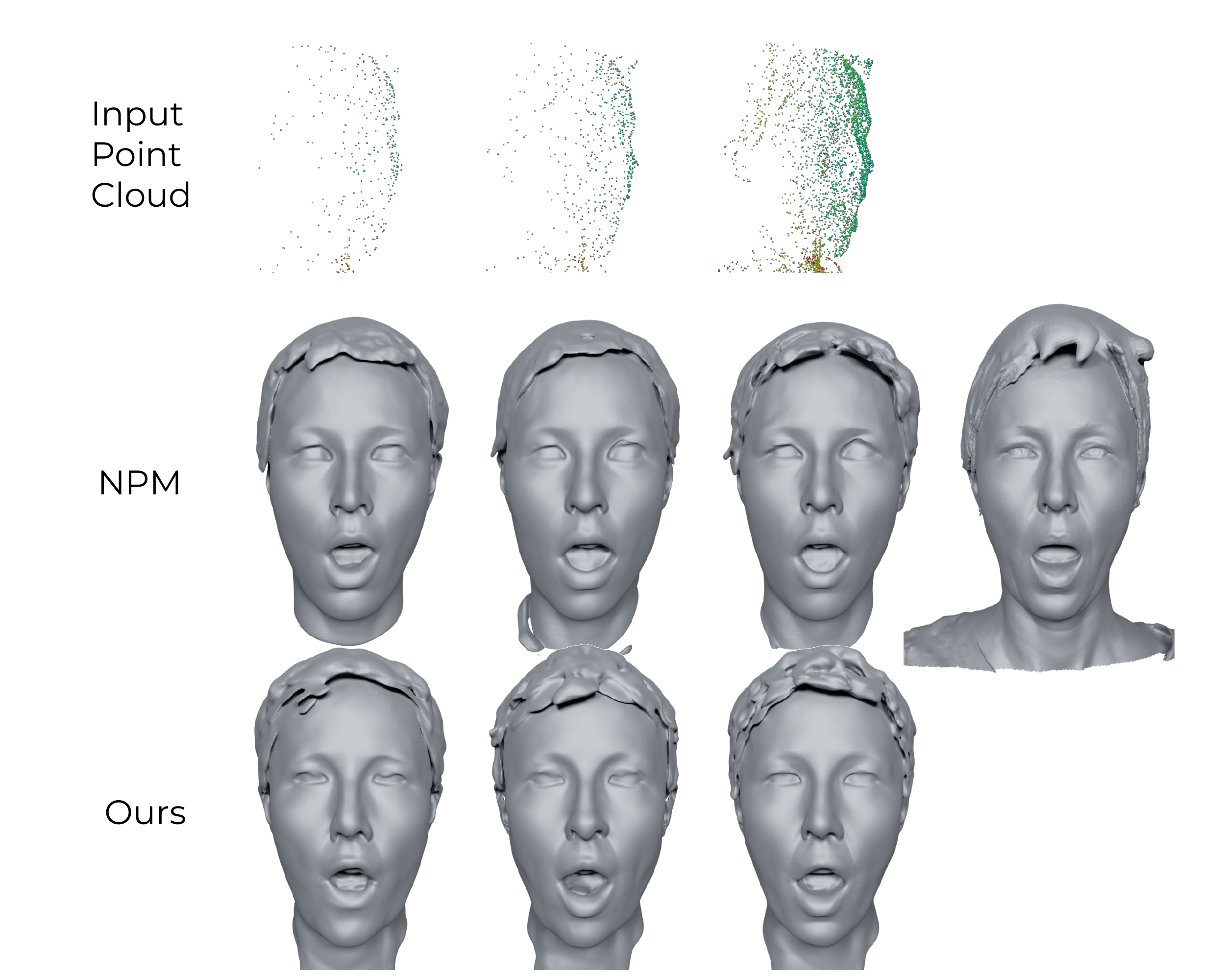}
    \begin{tabularx}{\linewidth}{P{0.16\linewidth}p{0.12\linewidth}P{0.15\linewidth}P{0.12\linewidth}P{0.18\linewidth}}
    \#points & 500 & 1000 & 5000 & GT Scan\\
    \end{tabularx}
    \caption{Qualitative comparison of NPMs~\cite{NPM} and our method with respect to the number of points in the input point cloud.}
    \label{fig:qual_num_points}
\end{figure}

\subsection{Deformation Consistency}

\begin{figure*}[tbh!]
    \centering
    \includegraphics[width=\linewidth]{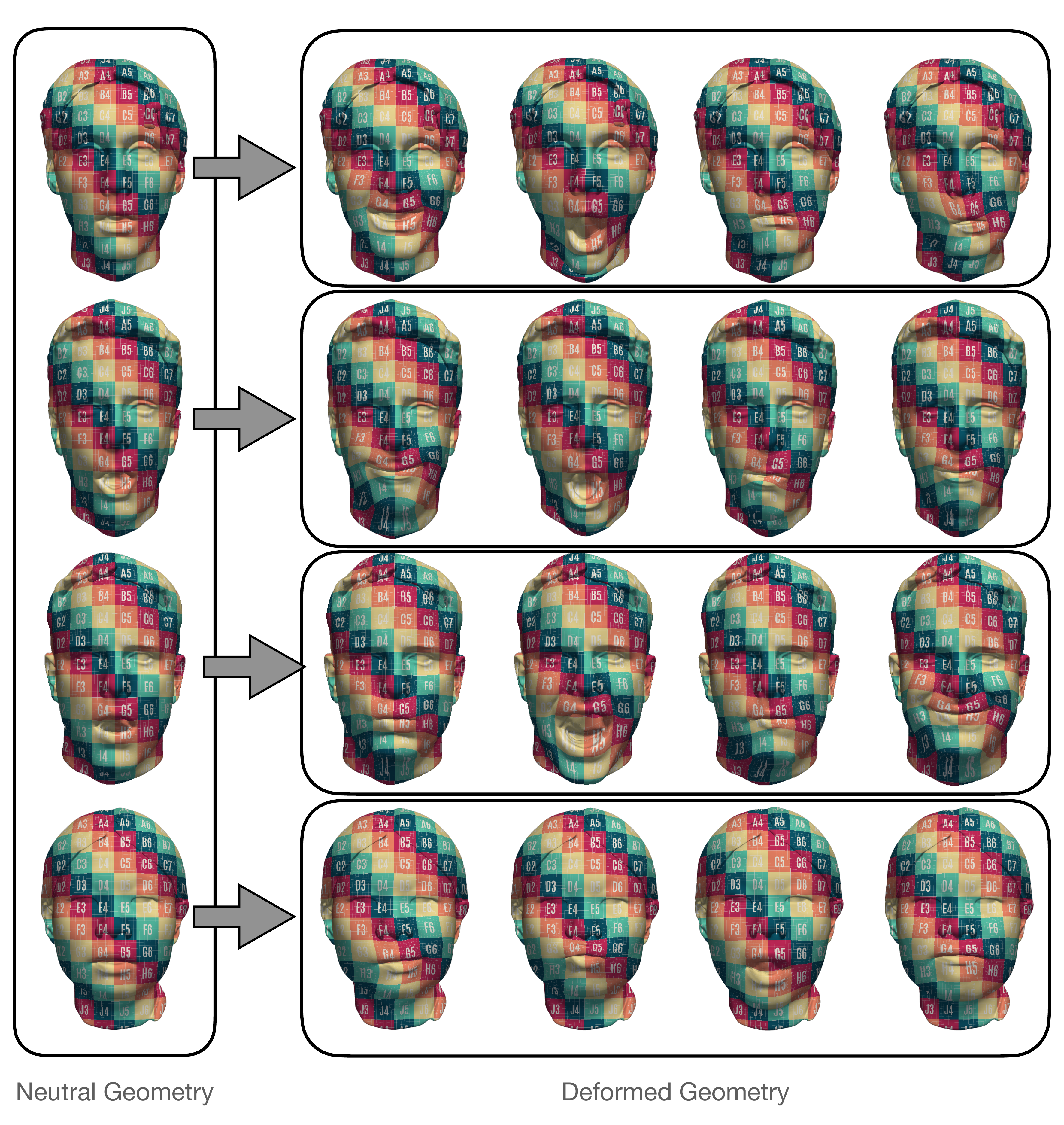}
    \caption{Deformation Consistency: We show surface correspondences between neutral and posed meshes from our test set. UV-coordinates are assigned to the mesh in canonical space after running marching cubes (left). The right side shows 4 different expressions for each example, which arise by deforming the neutral mesh, which preserved the uv-coordinates. }
    \label{fig:def_con}
\end{figure*}

Furthermore, we illustrate the behaviour of our expression network $\mathscr{F}_{\text{ex}}$ in figure \ref{fig:def_con}, by assigning a distinctive UV-map as colors to each vertex. To be more specific, we assign vertex colors by projecting a UV-map parallel to the "depth-dimension". We then fix vertex colors and deform the mesh using $\mathscr{F}_{\text{ex}}$. The results show that semantic consistency is preserved well, which is a direct consequence of our training strategy. Note that i3DMM~\cite{i3DMM} and ImFace~\cite{ImFace} report slightly less consistent correspondences.

\end{document}